\documentclass[lettersize,journal]{IEEEtran}

\usepackage{amsmath,amsfonts} 
\usepackage{array} 
\usepackage[caption=false,font=normalsize,labelfont=sf,textfont=sf]{subfig} 
\usepackage{textcomp} 
\usepackage{stfloats} 
\usepackage{url} 
\usepackage{verbatim} 
\usepackage{graphicx} 
\usepackage{cite} 
\usepackage{booktabs} 
\usepackage{multirow} 
\usepackage[table]{xcolor}
\usepackage[backref]{hyperref} 

\begin{document}


\title{OUS: Scene-Guided Dynamic Facial Expression Recognition}

\author{Xinji Mai, Haoran Wang, Zeng Tao, Junxiong Lin, Shaoqi Yan, Yan Wang*, Jing Liu, Jiawen Yu, Xuan Tong, Yating Li, and Wenqiang Zhang*,

\thanks{Xinji Mai, Haoran Wang, Zeng Tao, Junxiong Lin, Shaoqi Yan, Yan Wang, Jing Liu, Jiawen Yu and Xuan Tong are with the Academy for Engineering \& Technology, Fudan University, Shanghai, China. E-mail: see  \{xjmai23, hrwang23, ztao19, linjx23, sqyan19, yanwang19, jingliu19, jwyu23, xtong23\}@fudan.edu.cn }

\thanks{Wenqiang Zhang is with the Academy for Engineering \& Technology, Fudan University, Shanghai, China, and also with the School of Computer Science, Fudan University, Shanghai, China and the Yiwu Research Institute, Fudan University, Zhejiang. E-mail: see wqzhang@fudan.edu.cn}

\thanks{Manuscript received May 29, 2024;\\(*Corresponding author: Yan Wang and Wenqiang Zhang.)}

}

\markboth{Journal of \LaTeX\ Class Files,~Vol.~14, No.~8, August~2021}%
{Shell \MakeLowercase{\textit{et al.}}: A Sample Article Using IEEEtran.cls for IEEE Journals}


\maketitle

\begin{abstract}
Dynamic Facial Expression Recognition (DFER) is crucial for affective computing but often overlooks the impact of scene context. We have identified a significant issue in current DFER tasks: human annotators typically integrate emotions from various angles, including environmental cues and body language, whereas existing DFER methods tend to consider the scene as noise that needs to be filtered out, focusing solely on facial information. We refer to this as the Rigid Cognitive Problem. The Rigid Cognitive Problem can lead to discrepancies between the cognition of annotators and models in some samples. To align more closely with the human cognitive paradigm of emotions, we propose an Overall Understanding of the Scene DFER method (OUS). OUS effectively integrates scene and facial features, combining scene-specific emotional knowledge for DFER. Extensive experiments on the two largest datasets in the DFER field, DFEW and FERV39k, demonstrate that OUS significantly outperforms existing methods. By analyzing the Rigid Cognitive Problem, OUS successfully understands the complex relationship between scene context and emotional expression, closely aligning with human emotional understanding in real-world scenarios.
\end{abstract}



\section{Introduction}

\begin{figure}[t]
    \centering
    \includegraphics[width=3.5in]{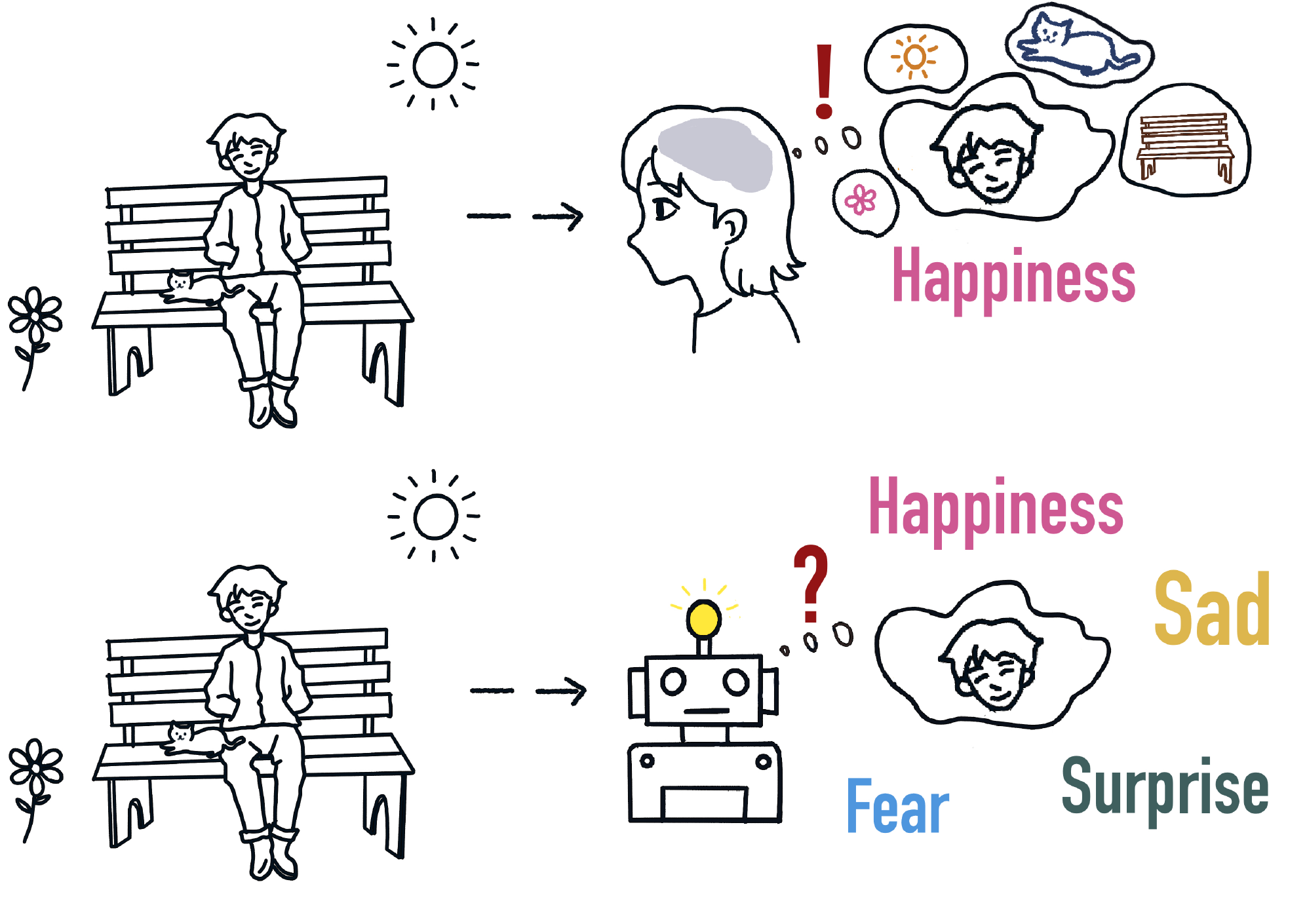}  
    \caption{Human annotators instinctively combine scene information and faces when labeling facial emotions. The DFER method only uses face information for emotion prediction.}
    \label{fig:fig1}
\end{figure}

\IEEEPARstart{D}{ynamic} Facial Expression Recognition is a crucial subfield of affective computing, facing a persistent challenge of ambiguous emotion classification. Typically, dynamic facial expressions that cannot be clearly classified by human annotators are removed during dataset construction. However, DFER methods still encounter a significant amount of ambiguous emotion classification. Through analyzing the performance of multiple DFER methods across different datasets, we found that these ambiguous classifications persist due to the discrepancy between annotators' and models' cognition, rather than actual annotation errors.

\begin{figure}[htbp]
    \centering
    \includegraphics[width=3in]{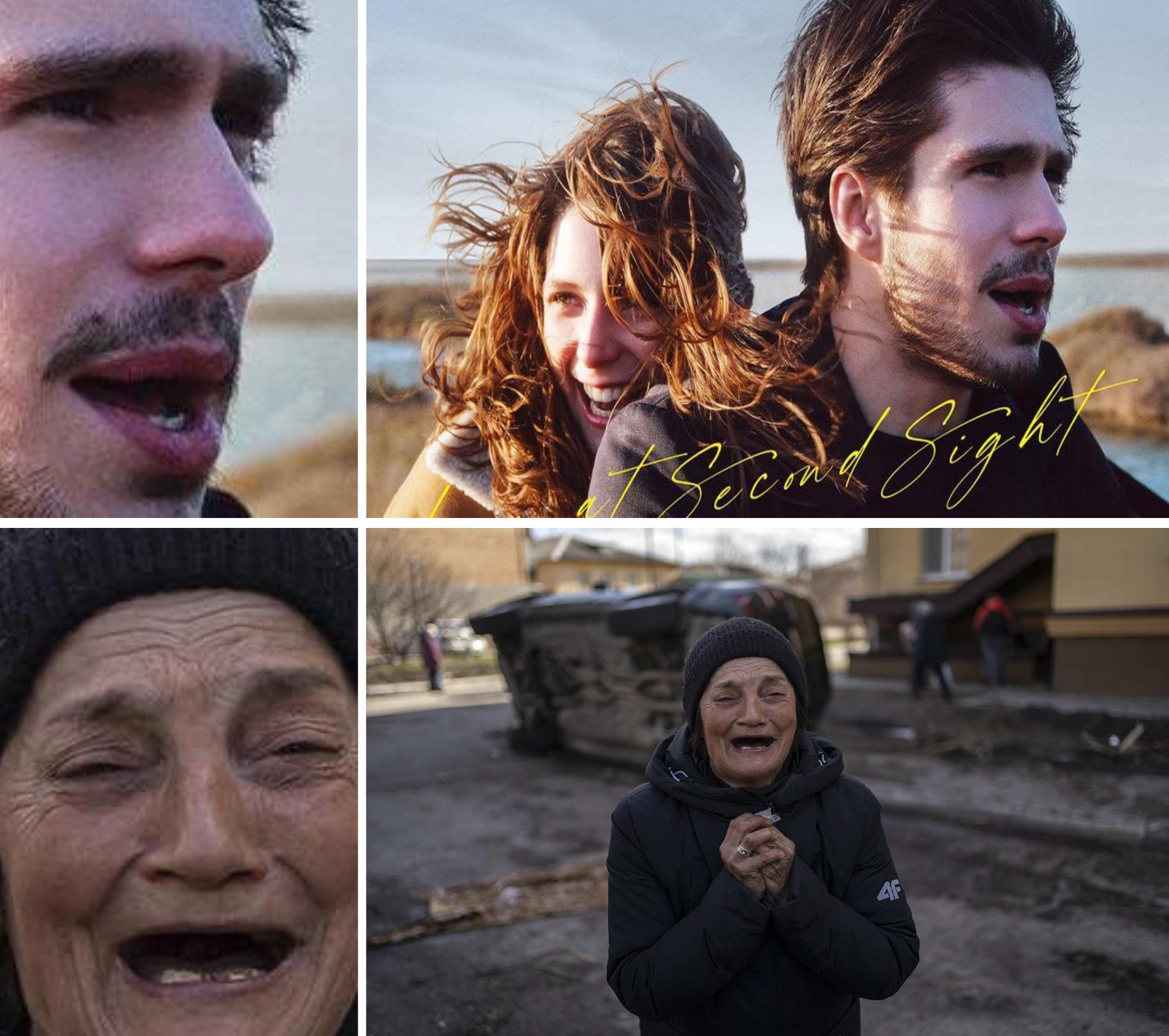}  
    \caption{Scene polarity helps determine mood. The left side shows facial expressions. When look at the face alone, the upper and lower expressions may be surprise, happiness, fear, sadness, etc., but by injecting the polarity of the scene, you can judge that these are two completely different expressions (happiness and sadness).}
    \label{fig:cont_happy}
\end{figure}

In DFER methods, almost all approaches unanimously consider scene information as noise and retain only the facial inputs. However, human annotators evaluate videos that include complete scene information when annotating facial expressions \cite{de2015perception}\cite{sinke2013body}. This discrepancy in the amount of information leads to what we call the Rigid Cognitive Problem. As illustrated in Fig. \ref{fig:cont_happy}, an example of the Rigid Cognitive Problem is presented. When viewing the face alone, the expression might convey sadness, fear, or other ambiguous emotions. However, with the scene context, human annotators can clearly identify it as a feeling of happiness and comfort in the figure above, and understand that the expression in the figure below is not joy but fear and sadness. In this instance, the positive scene polarity helps human annotators determine the ambiguous human expression more accurately \cite{righart2008recognition}\cite{porter2003blinded}\cite{sabatinelli2011emotional}. Therefore, we can define the Rigid Cognitive Problem as this, as shown in Fig. \ref{fig:fig1}. The Rigid Cognitive Problem refers to the human annotator instinctively combining scene information and facial information when annotating, and classifying unclearly faces are completed and preserved through scene information. For the DFER method, when using the data set, only face information is used for emotion classification, which leads to a decrease in performance. As shown in the Fig. \ref{fig:fig1} above, when analyzing emotions, human beings comprehensively consider the face, scene polarity, flowers, and sun and other objects. It is concluded that the conclusion of Happiness, and the model only obtains face information when analyzing emotions, so it is difficult to distinguish the unclear expression.

In psychology, some theories explain why the Rigid Cognitive PROBLEM occurs. Ekman \cite{ekman1971universals} pointed out that while certain basic facial expressions are universally recognized across cultures, accurately interpreting these expressions requires understanding the specific context. For example, a smile can signify happiness, embarrassment, or even anger, depending on the context. Schachter and Singer's Two-Factor Theory \cite{schachter1962cognitive}\cite{muller2013body} further explained that emotion results from the interaction between physiological arousal and its cognitive interpretation, influenced by the immediate environment and context. These theories demonstrate that when interpreting emotions, people consider various factors, including facial expressions, body language, and scene context, as do human annotators \cite{abramson2021social}\cite{metallinou2011tracking}.
Further interpretation of the Rigid Cognitive Problem reveals how scene information influences our judgment of ambiguous expressions. Environmental psychologist Roger Barker's Behavior Settings Theory \cite{hall1969ecological} emphasized that individuals' emotions and behaviors are significantly influenced by their surroundings. The emotional atmosphere of a lively amusement park versus a bloodshed refugee camp could lead to vastly different emotional responses. Scene polarity itself guides our judgment of ambiguous expressions, suggesting that positive scenes are more likely to be associated with positive emotions, and negative scenes with negative emotions \cite{duncan1969nonverbal}\cite{ wiener1972nonverbal}\cite{lafrance1978cultural}. Therefore, incorporating scene polarity into the variables considered by methods is an intuitive inference.

However, existing DFER methods typically focus on analyzing facial features in images while discarding other parts as noise, overlooking the subtle emotional and atmospheric nuances conveyed by these images, including scene polarity. For humans, integrating cues from scene polarity, scene objects, and facial expressions helps infer emotions, combining ambiguous facial expressions with broader scene context \cite{ziemke2007body}\cite{beck2010towards}. DFER methods lack this capability, with most focusing solely on facial classification, which is insufficient.

Recognizing the Rigid Cognitive Problem, we propose a new framework called Overall Understanding of the Scene to address this issue. Specifically, we aim to: First, 	isolate scene polarity classification from scene information to guide model classification of ambiguous expressions, employing polarity loss in OUS to separate scene polarity. Second, align scene information itself with facial information, considering the strong connection between non-facial information (e.g., body movements, object information) and emotions. OUS uses similarity loss for this alignment. Third, design a novel fusion module to merge scene and facial features, as they lack temporal relationships and are more about spatial perspectives. Existing fusion methods mostly target temporal sequences. OUS's Type Fusion Encoder (TFE) accomplishes this. Finally, develop an effective feature classification method and loss function to handle noisy information due to scene inclusion, as simple module classification is no longer suitable. OUS utilizes contrastive loss and flexible prompt design for this purpose.

Our research contributions are summarized as follows:
\begin{itemize}
\item We identify and analyze the Rigid Cognitive Problem, designing a dynamic facial expression recognition method called Overall Understanding of the Scene, which effectively aligns and integrates scene information to extract emotional knowledge from complex features. Extensive experiments on the two largest DFER datasets, DFEW and FERV39k, demonstrate that OUS significantly outperforms existing methods.
\item We design three loss functions—similarity loss, polarity loss, and contrastive loss—to constrain the training process, helping OUS align scene information, extract scene polarity, and classify emotional information. These multi-loss constraints effectively reduce the latent space distance between scene and human features, aiding in emotion classification through polarity guidance and contrastive loss.
\item We have designed a Type fusion encoder (TFE) that mainly uses cross-type attention mechanism to effectively integrate scenarios and facial information to integrate the characteristics related to emotion.
\end{itemize}

\begin{figure*}[h]
    \centering
    \includegraphics[width=7.5in]{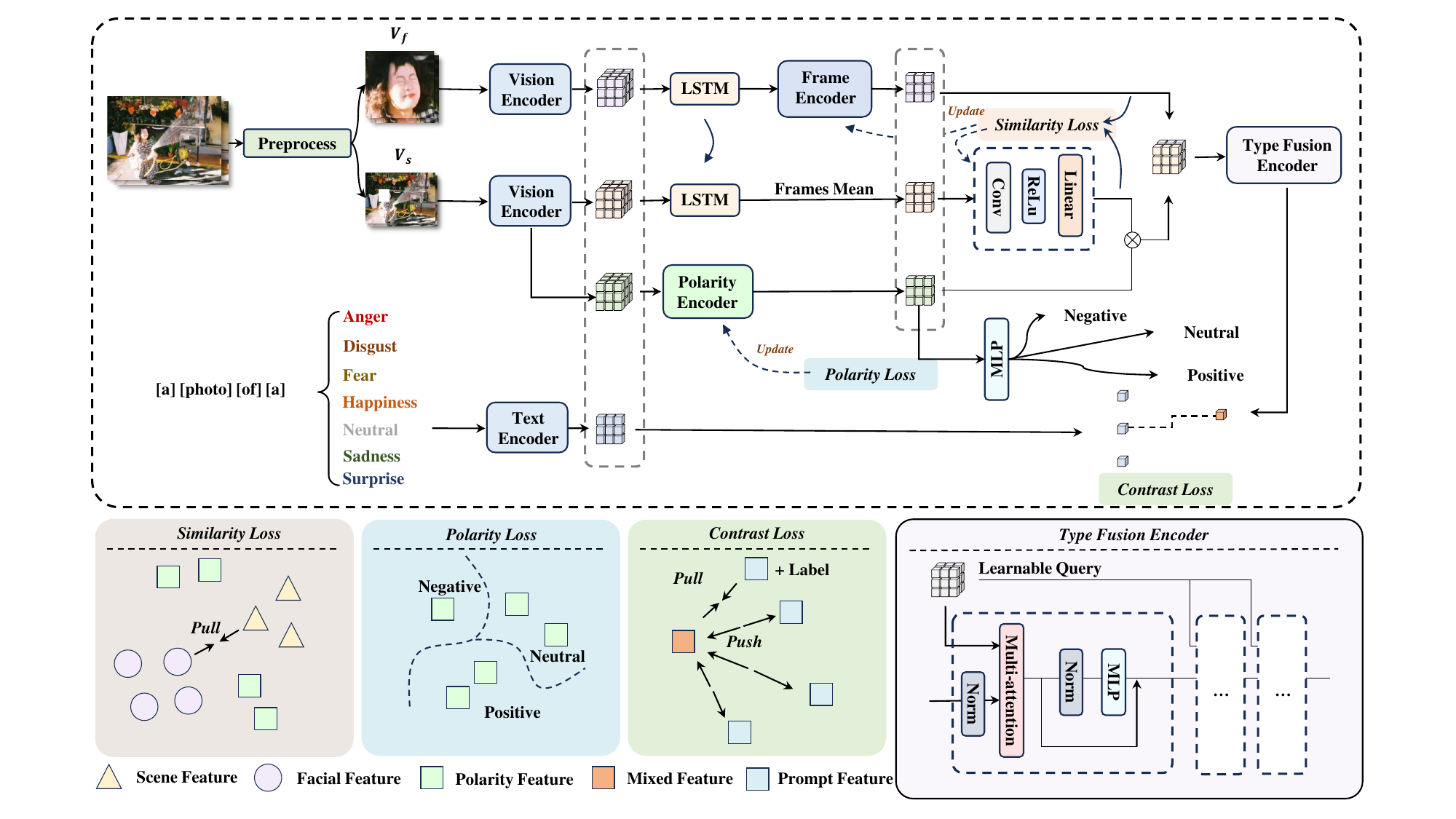}  

    \caption{{\bf{OUS Overall Architecture Diagram. }}OUS employs a dual-stream structure, separating images into facial and scene images through preprocessing. These images are encoded into latent spaces $V_f$ and $V_s$ using a shared Vision Encoder $f_V(\cdot)$. Temporal relationships are initially fused via LSTM. Facial features $V_f$ are further processed through a Frame Encoder, while scene features $V_s$, having less temporal variation, are processed through Frames Mean. Similarity Loss $L_{similarity}$ aligns the latent spaces of facial and scene features, reducing the distance between them using convolutional and fully connected layers. We also consider that the first four encoding blocks capture information like texture, color, and atmosphere, which are abstractly related to emotions. Polarity Loss constrains the Polarity Encoder to extract this polarity information, which is then concatenated with $V_{ft}$ and $V_{st}$ and fed into the Type Fusion Encoder. This mechanism uses Learnable Queries ($Q_{learn}$) as the Query, with $V_{ft}$ and $V_{st}$ as the Key and Value. Learnable Queries, initialized with a Gaussian distribution, are shared across all cross-type attention blocks and refined during training to capture emotional entities in the features. Finally, OUS incorporates substantial scene information, making fixed prompt methods unsuitable. Instead, it uses updatable prompts to compute contrastive loss with the output features.}
    \label{fig:pdfimage}
\end{figure*}

\section{RELATED WORK}
\subsection{DFER Methods}

Compared to static expression recognition, DFER is often more reliable due to the temporal correlations between different frames in facial sequences \cite{saleem2021real}\cite{guo2012dynamic}. In recent years, a series of advanced techniques and datasets have driven the development of this field\cite{fang2014facial}\cite{montirosso2010development}. These advancements provide tools for deeper understanding and analysis of facial expressions, paving the way for future applications \cite{liu2014learning}. 
In terms of data processing in the early stage, the work of Tao \cite{tao2023freq} pointed out that we can obtain high dynamic emotional fragments through the frequency based on a complete or complex videos, which lays the foundation for subsequent work. Recently, Vision Transformer (ViT) \cite{dosovitskiy2020image} based on the Transformer architecture \cite{vaswani2017attention} has shown remarkable potential in video emotion recognition, thanks to its superior feature extraction capabilities. Transformers, particularly their self-attention mechanisms, open new avenues for processing temporal sequence data and capturing long-range temporal dependencies. Originally designed for natural language processing, Transformers' ability to handle sequence data makes them perform exceptionally in DFER research. Besides traditional deep learning methods, approaches like CLIP \cite{radford2021learning}, CLIPER \cite{li2023cliper} and  $A^3$lign-DFER\cite{tao20243} have also gained significant attention in this field. CLIP, through a contrastive learning approach on large-scale image and text data, achieves powerful cross-modal representation capabilities. These methods significantly improve emotion recognition accuracy in natural and uncontrolled environments by integrating visual and textual information.
Our method differs from the aforementioned approaches that focus solely on extracting facial information from videos for emotion recognition. We emphasize recognizing complex features of the entire scene and the face within the video, treating scene information not as noise but as a valuable supplement for emotion understanding, aligning more closely with human emotional perception patterns.

\subsection{Scene Information in DFER Datasets}
The development and application of a series of datasets in the DFER field have greatly promoted its growth. CK+ \cite{lucey2010extended} is a classic facial expression dataset providing rich expression sequences, serving as a benchmark for evaluating various DFER algorithms. RAF-DB \cite{gera2021landmark} includes real-world facial expressions collected from the internet, highlighting the importance of emotion recognition in natural environments. AFEW \cite{ekenel2012benchmarking}, sourced from movies and TV shows, focuses on recognizing facial expressions in dynamic and realistic contexts. MMI \cite{valstar2010induced}, a multimodal facial expression database with controlled and spontaneous expressions, is particularly suited for analyzing subtle facial expression changes. DFEW \cite{jiang2020dfew}, as a more challenging dataset, contains facial expressions captured in natural environments, suitable for testing algorithms in real-world scenarios. FERV39k \cite{wang2022ferv39k} and other datasets extend the exploration range in the DFER field by providing a large number of facial expression samples with various attributes \cite{guo2016dynamic}.

Among these datasets, DFEW and FERV39k are the largest unconstrained DFER datasets, with sample sizes and difficulty levels far surpassing others. It is noteworthy that regardless of the dataset used in DFER tasks, they typically contain complete scene information rather than just facial details. The DFEW dataset contains a large number of real-life scenes, FERV39k contains data of up to 22 scenes, and other data sets contain a large number of scenes except human faces. However, most methods discard all scene information when recognizing facial information from these datasets. We argue that this approach is unrealistic, as scene information can provide valuable context for emotion understanding.


\section{Methodology}
In this section, we explore the complexity of the proposed OUS framework. The overall structure of OUS is shown in Fig. \ref{fig:pdfimage}. OUS mainly includes spatial encoding, temporal encoding, multi-Loss constraints and TFE fusion composition. Vision Encoder is used for spatial feature extraction. Frames Encoder is used to extract time characteristics. Similarity Loss is used to align the hidden space, and the Contrast Loss is used to classify emotional classification. TFE is used to integrate scene characteristics and facial features.

\subsection{Overview of Training Strategy for Different Losses}
Our framework primarily relies on a training strategy reinforced by three distinct loss functions to guide the optimization strategy. We posit that the introduction of scene information comprises three parts: the guidance of ambiance and texture information, the alignment with facial information in the latent space, and a loss that can flexibly match environmental entities. Specifically, color, textural, and luminance information, as well as related visual entities in the scene (such as carousels, artillery fire, etc.), can preliminarily determine the emotional polarity (positive, neutral, negative) of humans in that scene. For instance, in bright, flowery environments like amusement parks, the emotional polarity is likely positive, whereas in dark, bloody environments with ongoing artillery fire, it is more likely negative. We believe that this emotional polarity can guide the extraction of concepts related to emotions from the scene.

We use the polarity loss to optimize the polarity encoder so it can extract emotion-related polarity information from the output of the first four layers of the vision encoder to guide the cross-type attention mechanism. The polarity loss is defined as the cross-entropy loss:
\begin{equation}
L_{polarity} = -\sum_{i=1}^{N} y_i \log(\hat{y}_i),
\end{equation}
where $y_i$ is the true polarity label and $\hat{y}_i$ is the predicted polarity.

To fuse scene features with facial features, we need to align them in the same latent space. The similarity loss is used to optimize the frame encoder, linear, and convolutional layers, helping align scene features $V_s$ and facial features $V_f$ in the latent space and narrowing the relationship between scene features and emotion classification. The similarity loss is defined as the cosine similarity:
\begin{equation}
L_{similarity} = 1 - \frac{\sum_{i=1}^{N} (V_{f_i} \cdot V_{s_i})}{\sqrt{\sum_{i=1}^{N} V_{f_i}^2} \sqrt{\sum_{i=1}^{N} V_{s_i}^2}},
\end{equation}
where $V_{f_i}$ and $V_{s_i}$ represent the facial and scene features, respectively.

Finally, Due to the introduction of complex scene information, fixed prompts such as "a photo of" are no longer suitable for complex environmental scenes. Based on this, we use variable prompts that update during the training process. We compare the output features with the prompts using a contrastive loss function $L_{\text{cont}}$ to optimize the overall model, defined as:
\begin{equation}
L_{contrast} = -\sum_{i=1}^{N} \log \frac{\exp(\text{sim}(V_{s_i}, V_{f_i})/\tau)}{\sum_{j=1}^{N} \exp(\text{sim}(V_{s_i}, V_{f_j})/\tau)},
\end{equation}
where $\text{sim}$ denotes the similarity function, $\tau$ is a temperature scaling parameter, and $N$ is the batch size.

When the loss value exceeds $\alpha$, we employ a global loss, which is the sum of similarity loss, polarity loss, and contrast loss:
\begin{equation}
L_{global} = L_{similarity} + L_{polarity} + L_{contrast},
\end{equation}
and when it is less than $\alpha$, we only use the contrastive loss. This design helps to reduce the potential spatial distance between the scene and human features, and helps the model classify emotions effectively through polarity guidance and contrast loss. Besides, It helps our model converge quickly and find an appropriate solution space.

\subsection{Type Fusion Encoder}
In the fusion of facial and scene features, we designed a cross-type attention mechanism based on the attention mechanism. Unlike the block structure in transformers where attention is followed by normalization and then a feed-forward network, we first perform layer normalization on the facial and scene features used as keys and values. This is because we believe these features need to be normalized before extraction. 

\begin{equation}
V^{norm} = \text{LayerNorm}(V)
\end{equation}

Following this, when inputting into the multi-head attention mechanism, we use a shared Learnable Query ($Q{ learn} $), initialized with a Gaussian distribution, as the Query, with facial features $V_f^{norm}$ and scene features $V_s^{norm}$ as the Key and Value respectively. $Q{ learn} $ is shared across all cross-type attention blocks. During training, as backpropagation progresses, $Q{ learn} $ gradually denoises and learns to capture entities related to emotions from the facial and scene features, and filters the features that need attention from the Value matrix.

Learnable Queries are crucial for capturing the specific emotional entities within the context of the scene and facial expressions. A learnable query as it gradually denoises and learns, it pays attention to the parts related to emotion, and filters out the parts that do not need to be noticed in the process of multiplying with the value matrix, leaving only the features that need attention. These are initialized with a Gaussian distribution $\mathcal{N}(\mu, \sigma^2)$ to introduce variability and flexibility in capturing diverse features. The parameters of the Gaussian distribution are chosen to reflect the expected distribution of the features:

\begin{equation}
Q_{learn} \sim \mathcal{N}(\mu, \sigma^2)
\end{equation}

\noindent where $\mu$ and $\sigma^2$ are the mean and variance of the Gaussian distribution, respectively. This initialization ensures that the model starts with a diverse set of queries that can adapt to various features during training.

The output of the attention mechanism is then subjected to layer normalization:
\begin{equation}
O_{attn} = \text{LayerNorm}(\text{Attention}(Q_{learn}, K, V))
\end{equation}

Next, the normalized output is passed through a Multi-Layer Perceptron (MLP) as the feed-forward network:
\begin{equation}
O_{ff} = \text{MLP}(O_{attn})
\end{equation}

The MLP can be defined as:
\begin{equation}
\text{MLP}(x) = W_2 \cdot \text{ReLU}(W_1 \cdot x + b_1) + b_2
\end{equation}

Finally, the output of the MLP is normalized again:
\begin{equation}
O_{final} = \text{LayerNorm}(O_{ff})
\end{equation}

This design helps OUS effectively integrate facial and scene features. The layer normalization before the multi-head attention ensures that both types of features are on a similar scale, allowing the attention mechanism to more effectively learn the important features related to emotions. The use of a shared Learnable Query enables the model to focus on relevant entities and improve the emotional understanding of the scene and facial expressions.

\begin{figure*}[ht]
    \centering
    \includegraphics[width=7.5in]{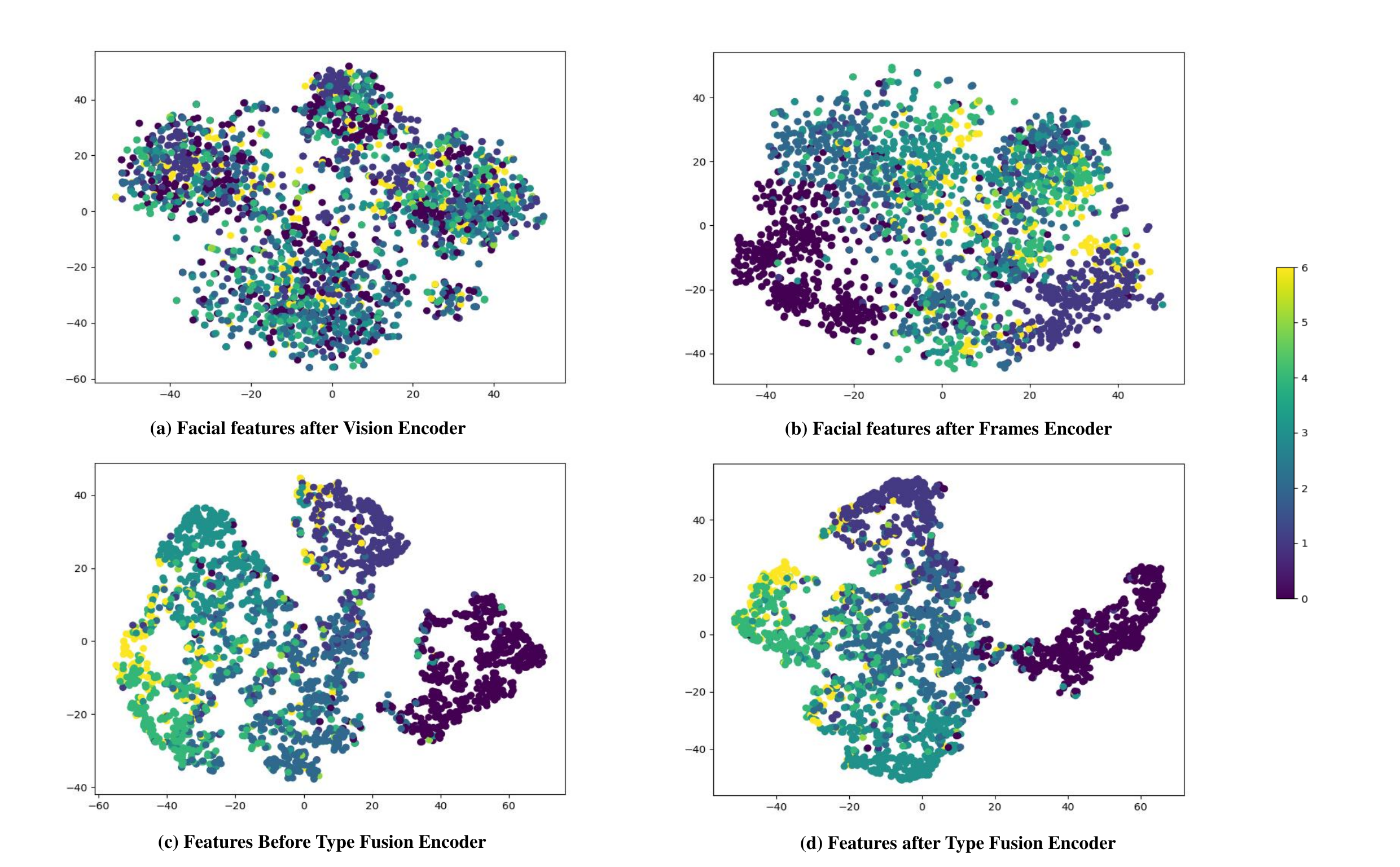}
    \caption{{\bf{Changes in Feature Clustering During the Inference Process.}} Global layout visualization of the feature space. In the legend, 0 to 6 represent happiness, sadness, neutral, anger, surprise, disgust, fear, respectively.}
    \label{fig:Changes_In_Feature}
\end{figure*}

\subsection{Prompt Engineering}
The application of prompt engineering in our OUS is inspired by the methodologies of the CLIP and CoOp papers. In CLIP, a series of text prompts are trained to assist the model in understanding and categorizing image content, while CoOp further learns a series of continuous vectors (i.e., learnable prompts) for zero-shot or few-shot learning on a pre-trained CLIP model. In our model, learnable prompts are used to guide the model in associating image features with emotional states. Specifically, we define a set of learnable vectors as prompts that are combined with image features to compute the final emotion classification probabilities. These prompts undergo optimization along with image features during training, capturing important semantic connections in the emotion recognition task. Specifically, the prompts given to the text encoder are designed as follows:

\begin{equation}
Prompt = [V]_{1}[V]_{2}\ldots[V]_{M} + [\text{CLASS}],
\end{equation}

\noindent Where each $[V]_m$ ($m \in \{1,\ldots, M\}$) is a vector of the same dimension as the word embeddings (i.e., 512 for CLIP ViT-B/32, 768 for CLIP ViT-L/14), and $M$ is a hyperparameter specifying the number of context tokens \cite{zhou2022conditional}.

In sum, prompt engineering provides our model with a flexible and powerful approach to understanding the complex relationships between images and textual prompts.

\subsection{Other detail of Network Architecture}
Video segments $\{V \mid V \in \mathbb{R}^{B \times T \times C \times H \times W}\}$ are initially processed by a preprocessing block $B_p(\cdot)$, resulting in the facial video sequence and the scene video sequence. Here, $B$, $T$, $C$, $H$, and $W$ represent the batch size, number of frames, channels, height, and width of the video sequence, respectively. The preprocessing module includes a crucial component: the facial recognition module, which separates facial and environmental information within the video. The facial recognition module is set to a frozen state, meaning its weights remain unchanged during the training process, ensuring stability and consistency from video input to feature extraction.

Subsequently, we encode the facial and environmental information into latent space facial features $V_f$ and scene features $V_s$ using a shared Vision Encoder. The weights we use are from ViT-L/14. Each input frame of the dual-stream facial features and scene features is processed independently. Frames are divided into $N$ patches and then flattened into a $D$-dimensional latent space as follows:
\begin{equation}
V_p = [v^1_pE; v^2_pE; \dots; v^N_pE] + E_{pos},
\end{equation}
where $v^i_p$ is the flattened patch vector, $E$ is the embedding matrix, and $E_{pos}$ is the positional embedding matrix.

The temporal ($T$) and batch ($B$) dimensions are merged into one dimension to form the facial features $V_f$ and the scene features $V_s$, such that $V \in \mathbb{R}^{B \cdot T \times N \times D}$ and $V \in \mathbb{R}^{B \cdot T \times N \times F}$, with $F$ denoting the feature dimension.

The Vision Encoder remains frozen throughout. Next, we input the facial features $V_f$ and scene features $V_s$ into an LSTM for early feature fusion. The LSTM processes the input features as follows:
\begin{equation}
h = \sigma(W_h \cdot [h_{\text{prev}}, x] + b_h),
\end{equation}
where $h$ represents the hidden state, $h_{\text{prev}}$ is the previous hidden state, $x$ is the input feature vector (either $V_f$ or $V_s$), and $\sigma$ is the activation function.

The facial temporal features are then processed to be $V_{ft}$ by a trainable Frames Encoder $f_f(\cdot)$, designed to capture the dynamic nature of facial expressions over time. We first re-transform the facial features back into tensor shape $B \times T \times F$ and then input them into the Frames Encoder to be $V_{st}$. Scene information, having less temporal variation, is extracted by computing the frames means:
\begin{equation}
V_{st} = \frac{1}{T} \sum_{t=1}^T V_s(t),
\end{equation}
providing a stable representation of the scene.

We use convolutional layers and fully connected layers to align the latent spaces of facial $V_{ft}$ and scene features $V_{st}$, reducing the distance between them. Similarity Loss $L_{sim}$ is used to update the Frame Encoder, convolutional layers, and fully connected layers. 

Additionally, the features output from the initial attention blocks during encoding contain information related to color, texture, and overall ambiance of the environment, which are abstractly linked to emotions. We use a Polarity Encoder to encode these features to be $V_{pol}$ and Polarity Loss $L_{pol}$ to update the Polarity Encoder. 

The output polarity features $V_{pol}$ are concatenated with the processed facial $V_{ft}$ and scene features $V_{st}$ and fed into the cross-type attention mechanism. The cross-type attention mechanism uses Learnable Queries $Q_{learn}$ as the Query, with facial and scene features as the Key and Value. The Learnable Queries $Q_{learn}$ are initially randomly initialized to a Gaussian distribution and shared across all cross-type attention blocks. During training, as backpropagation updates proceed, the Learnable Queries $Q_{learn}$ gradually denoise and learn to capture entities related to emotions in the $V_{pol}, V_{st}$, and $V_{ft}$. The attention mechanism is defined as:
\begin{equation}
\text{Attention}(Q, K, V) = \text{softmax}\left(\frac{Q_{learn}K^T}{\sqrt{d_k}}\right)V,
\end{equation}
where $Q$, $K$, and $V$ represent the Query, Key, and Value matrices, respectively, and $d_k$ is the scaling factor.

Finally, we believe that OUS introduces substantial scene information, making fixed prompt methods unsuitable. Instead, we use updatable prompts to compute contrastive loss $L_{contrast}$ with the output features.

\begin{table*}[ht]
\small
\renewcommand\arraystretch{1}
\centering
\setlength{\tabcolsep}{2.6mm}{
\caption{{\bf{Overall Model Performance Comparison}} (OUS vs. other SOTA methods on the DFEW dataset for seven-class classification)}
\begin{tabular}{lc|lllllll|ll}
\toprule
\textbf{Method}     & {\textbf{Publication}} & \textbf{Happy}        & \textbf{Sad}          & \textbf{Neutral}      & \textbf{Angry}        & \textbf{Surprise}      & \textbf{Disgust}      & {\textbf{Fear}} & \textbf{UAR}   & \textbf{WAR}   \\ \midrule
VGG13+LSTM         & /                               & 76.89                 & 37.65                 & 58.04                 & 60.7                  & 43.70                 & 0.00                  & 19.73                              & 42.39          & 53.70          \\
C3D                & CVPR’15                               & 75.17                 & 39.49                 & 55.11                 & 62.49                 & 45.00                 & 1.38                  & 20.51                              & 42.74          & 53.54          \\
ResNet18+LSTM      & /                               & 83.56                 & 61.56                 & 68.27                 & 65.29                 & 51.26                 & 0.00                  & 29.34                              & 51.32          & 63.85          \\
ResNet18+GRU       & /                               & 82.87                 & 63.83                 & 65.06                 & 68.51                 & 52.00                 & 0.86                  & 30.14                              & 51.68          & 64.02          \\
I3D-RGB           & CVPR’17                               & 78.61                 & 44.19                 & 56.69                 & 55.87                 & 45.88                 & 2.07                  & 20.51                              & 43.4           & 54.27          \\
P3D                & ICCV’17                               & 74.85                 & 43.40                 & 54.18                 & 60.42                 & 50.99                 & 0.69                  & 23.28                              & 43.97          & 54.47          \\
R(2+1)D18          & CVPR’18                                & 79.67                 & 39.07                 & 57.66                 & 50.39                 & 48.26                 & \underline{3.45}                  & 21.06                              & 42.79          & 53.22          \\
3D R.18+Center Loss & /                               & 78.49                 & 44.30                 & 54.89                 & 58.40                 & 52.35                 & 0.69                  & 25.28                              & 44.91          & 55.48          \\
3D Resnet18        & CVPR’18                              & 76.32                 & 50.21                 & 64.18                 & 62.85                 & 47.52                 & 0.00                  & 24.56                              & 46.52          & 58.27          \\
EC-STFL            & MM’20                              & 79.18                 & 49.05                 & 57.85                 & 60.98                 & 46.15                 & 2.76                  & 21.51                              & 45.35          & 56.51          \\
Former-DFER       & MM’21                               & 84.05                 & 62.57                 & 67.52                 & 70.03                 & 56.43                 & \textbf{3.45}                  & \underline{31.78}                              & 53.69          & 65.70          \\
NR-DFERNet         & arXiv’22                               & \underline{88.47}                 & 64.84                 & 70.03                 & 75.09                 & 61.60                 & 0.00                  & 19.43                              & 54.21          & 68.19          \\
GCA+IAL            & C\&C23                               & 87.95                 & 67.21                 & 70.10                 & \underline{76.06}                 & 62.22                 & 0.00                  & 26.44                              & 55.71          & 69.24          \\
SW-FSCL           & AAAI’23                               & 88.35                 & \underline{68.52}                 & \underline{70.98}                 & \textbf{78.17}        & \underline{64.25}                 & 1.42                  & 28.66                              & \underline{57.25}          & \underline{70.81}          \\
\rowcolor{gray!40} \textbf{OUS (Ours)}               & /              & \textbf{92.40}        & \textbf{75.23}        & \textbf{76.03}        & 74.33                 & \textbf{65.98}                 & 0.00                     & \textbf{44.12}                     & \textbf{60.94} & \textbf{74.10} \\ \bottomrule
\end{tabular}
\label{tab:Overall_Model_Performance_Comparison}
}

\end{table*}


\section{Experiment}
We aim to meticulously evaluate the performance of OUS in DFER tasks, conducted across two mainstream datasets in DFER: FERV39k and DFEW, which cover a wide range of real-world scenarios. Experiments on these datasets not only demonstrate our method's superior performance but also highlight the contributions of each module through ablation studies.

\begin{table}[h]
\small
\renewcommand\arraystretch{1}
\centering
\setlength{\tabcolsep}{2.7mm}{
\caption{{\bf{Results of OUS on the two datasets DFEW and FERV39k}}}

\begin{tabular}{l|cccc}
\toprule
\multicolumn{1}{c|}{\textbf{Method}} & \multicolumn{2}{c}{\textbf{DFEW}} & \multicolumn{2}{c}{\textbf{FERV39k}} \\ \cline{2-5} \specialrule{0em}{1pt}{2pt}
\multicolumn{1}{c|}{} & \multicolumn{1}{c}{UAR} & \multicolumn{1}{c}{WAR} & \multicolumn{1}{c}{UAR} & \multicolumn{1}{c}{WAR} \\ \midrule
C3D                   & 42.74 & 53.54 & 22.68 & 31.69 \\
P3D                   & 43.97 & 54.47 & 23.20 & 33.39 \\
I3D-RGB               & 43.40 & 54.27 & 30.17 & 38.78 \\
3D ResNet18           & 46.52 & 58.27 & 26.67 & 37.57 \\
R(2+1)D18             & 42.79 & 53.22 & 31.55 & 41.28 \\
ResNet18-LSTM         & 51.32 & 63.85 & 30.92 & 42.95 \\
ResNet18-ViT          & 55.76 & 67.56 & 38.35 & 48.43 \\
EC-STFL               & 45.35 & 56.51 & - & - \\
Former-DFER           & 53.69 & 65.70 & 37.20 & 46.85 \\
NR-DFERNet            & 54.21 & 68.19 & 33.99 & 45.97 \\
DPCNet                & 57.11 & 66.32 & - & - \\
EST                   & 53.94 & 65.85 & - & - \\
LOGO-Former           & 54.21 & 66.98 & 38.22 & 48.13 \\
IAL                   & 55.71 & 69.24 & 35.82 & 48.54 \\
CLIPER                & 57.56 & 70.84 & 41.23 & 51.34 \\
M3DFEL                & 56.10 & 69.25 & 35.94 & 47.67 \\
AEN                   & 56.66 & 69.37 & 38.18 & 47.88 \\
DFER-CLIP             & \underline{59.61} & \underline{71.25} & \underline{41.27} & \underline{51.65} \\
EmoCLIP               & 58.04 & 62.12 & 31.41 & 36.18 \\
\rowcolor{gray!40} \textbf{OUS (Ours)} & \textbf{60.94} & \textbf{74.10} & \textbf{42.43} & \textbf{53.30} \\ \bottomrule
\end{tabular}
\label{tab:Results_of_OUS}
}
\end{table}

\subsection{Training Details}
OUS was trained in a computing environment with 4 NVIDIA GeForce RTX 3090 GPUs and an Intel(R) Xeon(R) Gold 5218R CPU @ 2.10GHz. The training, based on CLIP's open-source code, utilized the Adam optimizer. With an initial learning rate of 0.002 and a batch size of 16, the model was trained for 60 epochs. The learning rate was reduced to a third of its value whenever the loss on the validation set didn't decrease for five consecutive epochs, and training was considered converged when the rate fell below 1e-7. The model was deemed overfitting if the training accuracy exceeded 80\%. The final model saved was the one with the lowest loss on the validation set. The entire training process followed strict data preprocessing and augmentation protocols for reliability and reproducibility.

\begin{figure}[ht]
    \centering
    \includegraphics[width=3.5in]{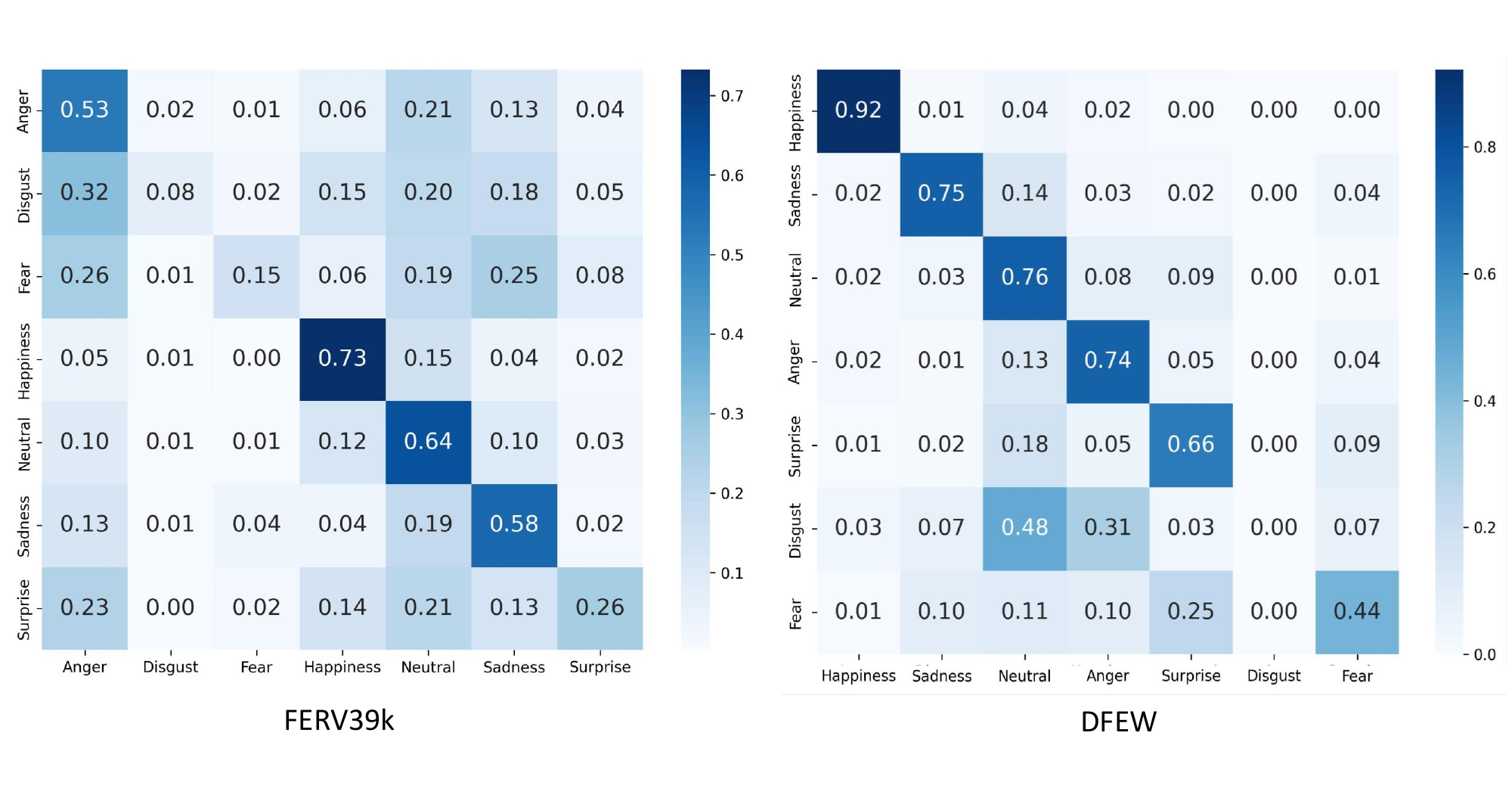}  
    \caption{{\bf{Confusion Matrices on the DFEW and FERV39k Dataset.}}}
    \label{fig:convergence}
\end{figure}

\subsection{Performance Evaluation}
OUS's performance was assessed on FERV39k and DFEW datasets, using weighted average recall (WAR) and unweighted average recall (UAR) as the primary metrics. Compared to existing advanced methods, our results outlined in Tables \ref{tab:Overall_Model_Performance_Comparison} and \ref{tab:Results_of_OUS} show OUS's superior performance. OUS has increased by 2.85\% on the DFEW dataset than the current SOTA method, and an increase of 1.65\% on the FERV39k dataset.

Observing Fig. \ref{fig:convergence}, it is apparent that across the two datasets, the categories of Happiness, Sad, Surprise and Fear exhibit the highest recognition accuracy. Obviously, Disgust and Fear are samples with the lowest accuracy of classification. This is also a characteristic of the DFER field. Due to the serious long tail distribution effect of data sets, there are fewer expressions of dislikes and fear. Furthermore, as per the results in Table \ref{tab:Overall_Model_Performance_Comparison}, these categories show the most significant improvement compared to other methods. This aligns with our initial hypothesis that considering contextual scene information can enhance recognition accuracy. The ease of acquiring and recognizing contextual cues for Happiness, Sad, Surprise and Fear, such as closely related body movements and either comfortable or tense environments, supports this assertion. Happiness, Sad, Surprise and Fear are easily confused expressions, which are difficult to classify in separate face recognition, but the classification accuracy increases greatly after injection of scene information, which also proves the existence of Rigid Cognitive Problem and the effective solution in OUS method.

In our comparative analysis, we meticulously reviewed DFER methods from the last decade, with a focus on the prevailing state-of-the-art methods. These include 3D Convolutional Neural Networks (C3D) \cite{tran2015learning}, Inflated 3D ConvNet (I3D-RGB) \cite{carreira2017quo}, Pseudo-3D Residual Networks (P3D) \cite{qiu2017learning}, and various configurations of 3D ResNet18 \cite{hara2018can}. We also evaluated methods based on the ResNet architecture, such as ResNet18 combined with LSTM networks \cite{he2016deep}, ResNet18 with Gated Recurrent Units (GRU), and ResNet18-ViT \cite{dosovitskiy2020image}. Additionally, our review covered CLIP-based methods, notably CLIPER \cite{li2023cliper}, DFER-CLIP \cite{zhao2023prompting}, EmoCLIP \cite{foteinopoulou2023emoclip}, and other approaches like Former-DFER \cite{jiang2020dfew}, LOGO-Former \cite{ma2023logo}, NR-DFERNet \cite{li2022nr}, T-ESFL \cite{li2023scaling}, DPCNet \cite{wang2022dpcnet}, IAL \cite{li2023intensity}, M3DFEL \cite{wang2023rethinking}, AEN \cite{lee2023frame}, SW-FSCL \cite{yan2023empower}, EC-STFL \cite{jiang2020dfew}, and others. Cumulatively, our method demonstrated the highest effectiveness on the FERV39k and DFEW datasets, showcasing its robustness in the dynamic facial expression recognition domain.

\begin{figure*}[ht]
    \centering
    \includegraphics[width=7.5in]{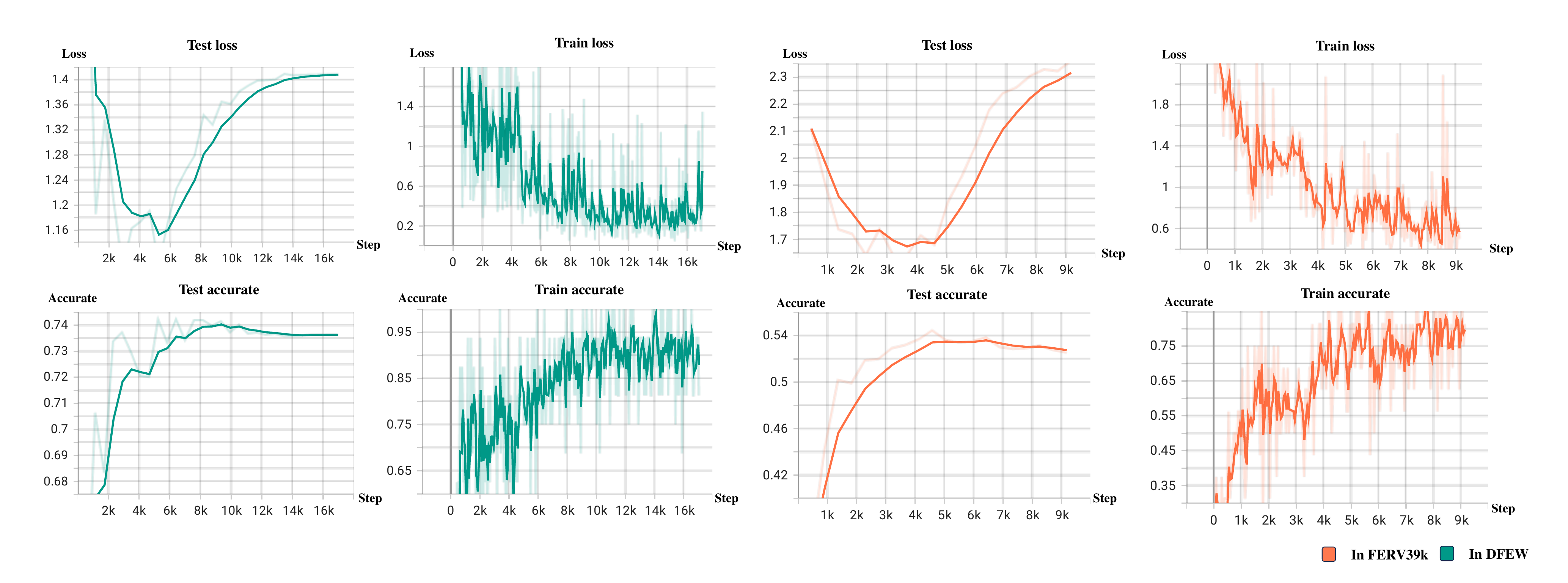}  
    \vspace{-7mm}
    \caption{{\bf{Loss and Accurate in the process of OUS training In DFEW and FERV39k Dataset.}}}
    \label{fig:Loss}
\end{figure*}

\begin{table}[htbp]
\renewcommand\arraystretch{1}
\centering
\setlength{\tabcolsep}{3.8mm}{
\caption{{\bf{Results of hyperparametric ablation with different experimental settings on DFEW Dataset}}}
\begin{tabular}{c|cc|cc}
\toprule
\multicolumn{1}{c|}{\textbf{Index}} & \multicolumn{2}{c|}{\textbf{Setting}} & \multicolumn{2}{c}{\textbf{Evaluation}} \\ \cline{2-5} \specialrule{0em}{1pt}{2pt}
 & TFE & Length & UAR & WAR \\ \midrule
1 & 8 & 16 & 0.56926 & 0.70598 \\
2 & 8 & 32 & 0.59908 & 0.72906 \\
3 & 8 & 64 & 0.60973 & 0.73675 \\
4 & 12 & 16 & 0.5842 & 0.7188 \\
5 & 12 & 32 & \textbf{0.6324} & 0.72393 \\
\rowcolor{gray!40} \textbf{6 (Ours)} & 12 & 64 & 0.60937 & \textbf{0.74103} \\
7 & 16 & 16 & 0.60145 & 0.73504 \\
8 & 16 & 32 & 0.59968 & 0.7359 \\
9 & 16 & 64 & 0.61615 & 0.74017 \\ \bottomrule
\end{tabular}
\label{tab:tfe_prompt}
}
\end{table}

\subsection{Ablation Study}
\noindent\textbf{Module ablation.} To ensure the fairness and rigor of our experiments, we performed ablation studies by selectively removing the multi-loss constraints and replacing the Type Fusion Encoder with average pooling of the scene and facial features. We compared three settings: using only the cross loss, using multi-loss constraints (similarity loss, polarity loss, and contrastive loss), and replacing the TFE with average pooling.

We first examined the impact of the multi-loss constraints. The results, shown in Table \ref{tab:multi_loss_ablation} and \ref{tab:multi_loss_ablation_dfew}, indicate that both the WAR and UAR significantly decreased when only the contrastive loss was used. This suggests that the interaction between the three losses is a crucial component of the model's effectiveness. When using only the cross loss, we observed a notable increase in convergence cycles and a decrease in performance. This may be because the solution space constrained by the three losses is easier to converge, highlighting the importance of the multi-loss strategy.

We investigated the impact of the TFE by replacing it with average pooling of the scene and facial features. The results, shown in Table \ref{tab:multi_loss_ablation} and \ref{tab:multi_loss_ablation_dfew}, reveal a significant decrease in accuracy when the TFE is removed. This underscores the TFE's role in capturing the emotional context from facial and scene features for emotion prediction. The absence of the TFE module resulted in a marked reduction in both UAR and WAR, indicating that the TFE is essential for extracting emotion-related connections between facial and scene features.

We also analyzed the convergence behavior of the model under different settings. Table \ref{tab:multi_loss_ablation} and \ref{tab:multi_loss_ablation_dfew} illustrates the convergence cycles for the cross loss and multi-loss settings. The model using only the cross loss exhibited significantly slower convergence and lower final performance, as evidenced by the increased number of epochs required to achieve optimal performance. This further emphasizes the effectiveness of the multi-loss strategy in facilitating faster convergence and better performance.

The ablation studies highlight the importance of the multi-loss constraints and the TFE in our OUS model. The significant drop in UAR and WAR when using only the cross loss or replacing the TFE with average pooling demonstrates that these components are integral to the model's ability to capture and predict emotions accurately. The multi-loss constraints provide a solution space that is easier to converge, thereby improving the model's performance. Similarly, the TFE plays a critical role in extracting emotion-related connections between facial and scene features, which are essential for accurate emotion prediction.

\begin{table*}[ht]
\renewcommand\arraystretch{1}
\small
\centering
\setlength{\tabcolsep}{2.4mm}{
\caption{{\bf{Comparative results for Loss and TFE in FERV39k dataset, $L_{global}$ and $L_{ce}$ represents the 3-loss constraints and Cross entropy.}}}
\begin{tabular}{l|lllllll|ll|c}
\toprule
\textbf{Strategy} & \textbf{Angry} & \textbf{Disgust} & \textbf{Fear} & \textbf{Happy} & \textbf{Neutral} & \textbf{Sad} & \textbf{Surprise} & \textbf{UAR} & \textbf{WAR} & \textbf{Best Epoch} \\ \midrule 
$L_{ce}$ & 40.01 & 8.12 & \textbf{20.07} & 71.13 & 51.78 & 49.02 & 17.99 & 36.65 & 45.31 & 15 \\
$L_{global}$  & 39.01 & \textbf{10.10} & 10.17 & \textbf{77.22} & \textbf{67.13} & 46.77 & 21.33 & 38.48  & 49.50 & \textbf{6} \\
$L_{ce}$ + TFE & 50.03 & 7.01 & 4.13 & 71.22 & 60.03 & \textbf{60.17} & \textbf{36.11} & 40.88 & 51.78 & 18 \\
\rowcolor{gray!40} $L_{global}$ + TFE & \textbf{53.21} & 8.13 & 15.42 & 73.01 & 64.45 & 58.10 & 25.98 & \textbf{42.43} & \textbf{53.30} & \textbf{6} \\ \bottomrule
\end{tabular}
\label{tab:multi_loss_ablation}
}
\end{table*}

\begin{table*}[htbp]
\renewcommand\arraystretch{1}
\small
\centering
\setlength{\tabcolsep}{2.4mm}{
\caption{{\bf{Comparative results for Loss and TFE in DFEW dataset, $L_{global}$ and $L_{ce}$ represents the 3-loss constraints and Cross entropy.}}}
\begin{tabular}{l|lllllll|ll|c}
\toprule
\textbf{Strategy} & \textbf{Happy} & \textbf{Sad} & \textbf{Neutral} & \textbf{Angry} & \textbf{Surprise} & \textbf{Disgust} & \textbf{Fear} & \textbf{UAR} & \textbf{WAR} & \textbf{Best Epoch} \\ \midrule 
$L_{ce}$ & 93.01 & \textbf{80.45} & \textbf{83.02} & 32.13 & 60.02 & 0.00 & 37.11 & 55.12 & 67.82 & 10 \\
$L_{global}$ & 92.44 &79.31 & 67.97 &68.42 &63.23 & 3.11 & 39.98 & 59.12  & 71.20 & 5 \\
$L_{ce}$ + TFE & 92.15 & 79.13 & 66.01 & \textbf{75.02} & 57.79 & \textbf{17.17} & 38.33 & 60.81 & 71.58 & 11 \\
\rowcolor{gray!40} $L_{global}$ + TFE &\textbf{92.40}  & 75.23   & 76.03 & 74.33  & \textbf{65.98}  & 0.00  & \textbf{44.12}  & \textbf{60.94} & \textbf{74.10} & \textbf{5} \\ \bottomrule
\end{tabular}
\label{tab:multi_loss_ablation_dfew}
}
\end{table*}

\begin{table*}[htbp]
\renewcommand\arraystretch{1}
\small
\centering
\setlength{\tabcolsep}{3.8mm}{
\caption{{\bf{Ablation Results of Different Prompt Length on DFEW Dataset.}}}
}
\begin{tabular}{c|ccccccc|cc}
\toprule
\multicolumn{1}{l|}{\textbf{Length}} & \multicolumn{1}{l}{\textbf{Happy}} & \multicolumn{1}{l}{\textbf{Sad}} & \multicolumn{1}{l}{\textbf{Neutral}} & \multicolumn{1}{l}{\textbf{Angry}} & \multicolumn{1}{l}{\textbf{Surprise}} & \multicolumn{1}{l}{\textbf{Disgust}} & \multicolumn{1}{l|}{\textbf{Fear}} & \multicolumn{1}{l}{\textbf{UAR}} & \multicolumn{1}{l}{\textbf{WAR}} \\ \midrule
16                                                                                     &\textbf{94.12}                     & 80.77                   &\textbf{82.69}                       & 62.01                     & 45.99                        & 0.00                        &\textbf{46.06}                     & 58.57                   & 72.13                   \\
32                                                                                     & 92.11                     & 79.42                   & 71.31                       & 76.2                      & 62.03                        & 0.00                        & 43.17                     & 60.50                   & 73.50                   \\
\rowcolor{gray!40} 64                                                                                     & 92.40        & 75.23        & 76.03        & \textbf{74.33}                 & \textbf{65.98}                 & \textbf{0.00}                     & 44.12                     & \textbf{60.94} & \textbf{74.10}                   \\ \bottomrule
\end{tabular}
\label{tab:prompt_length}
\end{table*}

\noindent\textbf{Hyperparametric ablation. }We performed extensive ablation studies to verify the superiority of our selected hyperparameters. The experiments are divided into two main parts: the investigation of prompt length and the combined ablation of TFE block numbers and prompt length.
﻿
We explored different prompt lengths to identify the optimal setting. Our findings indicate a positive correlation between prompt length and method performance. Longer prompts provide more flexibility and better capacity to model the additional scene information. The detailed results are presented in Table \ref{tab:prompt_length}, showing that a prompt length of 64 yields the best results, significantly improving the model's accuracy and robustness.
﻿
We also conducted combined ablation studies on the number of TFE blocks and the prompt length. The purpose was to determine the best combination that maximizes model performance. The results, detailed in Table \ref{tab:tfe_prompt}, demonstrate that the model performs optimally when the TFE block number is set to 12 and the prompt length is 64. This configuration effectively captures the nuanced interactions between facial and scene features, enhancing the overall emotion recognition accuracy.

\subsection{Discussion}

We compare the performance of OUS with other state-of-the-art (SOTA) methods on the DFEW dataset for seven-class classification. The results are presented in Table~\ref{tab:Overall_Model_Performance_Comparison}.
﻿
The performance of OUS on the DFEW and FERV39k datasets is summarized in Table~\ref{tab:Results_of_OUS}. We compare our method against several baseline models, highlighting the superior performance of OUS.
﻿
The results in Table~\ref{tab:Overall_Model_Performance_Comparison} reveal that the inclusion of scene information significantly improves the classification of happiness, sadness, and fear. This aligns with our initial hypothesis that these emotions have a strong correlation with the surrounding scene context and body movements.
﻿
Tables~\ref{tab:Overall_Model_Performance_Comparison} and \ref{tab:Results_of_OUS} provide evidence that our method outperforms existing state-of-the-art DFER approaches over the past decade. In the Happy classification, OUS is 4.05\% higher than the SOTA method, 6.71\% higher than the current SOTA method in Sad classification, 5.05\% higher than the current SOTA method than the current SOTA method in Neutral, and 1.73\% higher than the current SOTA method in Surprise. 12.35\% of the current SOTA method on Fear. All in all, OUS leads by 2.85\% on the DFEW dataset and 1.65\% on the FERV39k dataset compared to current SOTA methods. 
﻿
As illustrated in Table~\ref{tab:multi_loss_ablation} and Table~\ref{tab:prompt_length}, there is a positive correlation between the length of learnable prompts and model accuracy. It is noteworthy that models using only the cross loss function without the TFE module performed the worst. In contrast, our model with three loss constraints and the TFE module showed a performance improvement of 7.99\%, and the convergence period reduced from 15 epochs to 6 epochs. As Shown in Fig. \ref{fig:Loss}, whether on DFEW or FERV39k dataset, the Loss and Accurate of the train and test set are completely converged within 5K steps, that is, in the 6 epochs can reach approximate convergence, and the score all reach SOTA level, surpassing the current SOTA method. Furthermore, as shown in Fig. \ref{fig:Changes_In_Feature}, first look at the (a) figure. This is the feature distribution of the images of the DFEW and FERV39k data sets in the Vision Encoder. Because the Vision Encoder is frozen, the features are evenly distributed. The preliminary gathering phenomenon of features after entering Frame Encoder is preliminary in (b) figure. This is due to the optimization of the Similarity Loss, which makes the face characteristics gather. The features in the figure Due to the optimization of Multi-loss, the overall gathering is more obvious in (c) figure. (d) Show the powerful gathering effect of the Type Fusion Encoder we designed. The features are significantly clustered after passing through the TFE module, substantiating the substantial contribution of our three loss constraints and the TFE module to the effectiveness of OUS. This also confirms our hypothesis that the injection of polarity helps to classify emotions.

\section{Conclusion}
In this paper, we identified and analyzed the Rigid Cognitive Problem, a prevalent issue and methodological bias in DFER tasks. To address this problem, we designed a dynamic facial expression recognition method called Overall Understanding of the Scene (OUS), which effectively aligns and integrates scene information to extract emotional knowledge from complex features. The method primarily utilizes multiple loss constraints to reduce the latent space distance between scene and human features and efficiently classify emotions. Additionally, the Type Fusion Encoder (TFE) with a cross-type attention mechanism and learnable queries effectively integrates scene and facial information. Extensive experiments conducted on the DFEW and FERV39k datasets demonstrated significant improvements over existing methods, validating the robustness and effectiveness of our approach. Overall, We have discovered and analyzed the Rigid Cognitive Problem and proposed the OUS method to address this issue. The superior performance of OUS illustrates the importance of utilizing scene information in the DFER task, and we hope our work can inspire other researchers.

\section*{Acknowledgments}
This work was supported by National Key Research and Development Program of China (2023YFC3604802), China Postdoctoral Science Foundation under Grant (2023M730647, 2023TQ0075), by National Natural Science Foundation of China (No.62072112); and by Scientific and Technological Innovation Action Plan of Shanghai Science and Technology Committee (22511101502, 22511102202).

\bibliographystyle{IEEEtran}
\bibliography{mybibfile}

\begin{thebibliography}{10}
\providecommand{\url}[1]{#1}
\csname url@samestyle\endcsname
\providecommand{\newblock}{\relax}
\providecommand{\bibinfo}[2]{#2}
\providecommand{\BIBentrySTDinterwordspacing}{\spaceskip=0pt\relax}
\providecommand{\BIBentryALTinterwordstretchfactor}{4}
\providecommand{\BIBentryALTinterwordspacing}{\spaceskip=\fontdimen2\font plus
\BIBentryALTinterwordstretchfactor\fontdimen3\font minus \fontdimen4\font\relax}
\providecommand{\BIBforeignlanguage}[2]{{%
\expandafter\ifx\csname l@#1\endcsname\relax
\typeout{** WARNING: IEEEtran.bst: No hyphenation pattern has been}%
\typeout{** loaded for the language `#1'. Using the pattern for}%
\typeout{** the default language instead.}%
\else
\language=\csname l@#1\endcsname
\fi
#2}}
\providecommand{\BIBdecl}{\relax}
\BIBdecl

\bibitem{de2015perception}
B.~De~Gelder, A.~W. de~Borst, and R.~Watson, ``The perception of emotion in body expressions,'' \emph{Wiley Interdisciplinary Reviews: Cognitive Science}, vol.~6, no.~2, pp. 149--158, 2015.

\bibitem{sinke2013body}
C.~B. Sinke, M.~E. Kret, and B.~de~Gelder, ``Body language: Embodied perception of emotion,'' in \emph{Measurement With Persons}.\hskip 1em plus 0.5em minus 0.4em\relax Psychology Press, 2013, pp. 349--366.

\bibitem{righart2008recognition}
R.~Righart and B.~d. Gelder, ``Recognition of facial expressions is influenced by emotional scene gist,'' \emph{Cognitive, Affective, \& Behavioral Neuroscience}, vol.~8, no.~3, pp. 264--272, 2008.

\bibitem{porter2003blinded}
S.~Porter, L.~Spencer, and A.~R. Birt, ``Blinded by emotion? effect of the emotionality of a scene on susceptibility to false memories.'' \emph{Canadian Journal of Behavioural Science/Revue canadienne des sciences du comportement}, vol.~35, no.~3, p. 165, 2003.

\bibitem{sabatinelli2011emotional}
D.~Sabatinelli, E.~E. Fortune, Q.~Li, A.~Siddiqui, C.~Krafft, W.~T. Oliver, S.~Beck, and J.~Jeffries, ``Emotional perception: meta-analyses of face and natural scene processing,'' \emph{Neuroimage}, vol.~54, no.~3, pp. 2524--2533, 2011.

\bibitem{ekman1971universals}
P.~Ekman, ``Universals and cultural differences in facial expressions of emotion.'' in \emph{Nebraska symposium on motivation}.\hskip 1em plus 0.5em minus 0.4em\relax University of Nebraska Press, 1971.

\bibitem{schachter1962cognitive}
S.~Schachter and J.~Singer, ``Cognitive, social, and physiological determinants of emotional state.'' \emph{Psychological review}, vol.~69, no.~5, p. 379, 1962.

\bibitem{muller2013body}
C.~M{\"u}ller, A.~Cienki, E.~Fricke, S.~Ladewig, D.~McNeill, and S.~Tessendorf, \emph{Body-Language-Communication. Volume 1}.\hskip 1em plus 0.5em minus 0.4em\relax Walter de Gruyter, 2013.

\bibitem{abramson2021social}
L.~Abramson, R.~Petranker, I.~Marom, and H.~Aviezer, ``Social interaction context shapes emotion recognition through body language, not facial expressions.'' \emph{Emotion}, vol.~21, no.~3, p. 557, 2021.

\bibitem{metallinou2011tracking}
A.~Metallinou, A.~Katsamanis, Y.~Wang, and S.~Narayanan, ``Tracking changes in continuous emotion states using body language and prosodic cues,'' in \emph{2011 IEEE international conference on acoustics, speech and signal processing (ICASSP)}.\hskip 1em plus 0.5em minus 0.4em\relax IEEE, 2011, pp. 2288--2291.

\bibitem{hall1969ecological}
E.~T. Hall, ``Ecological psychology: Concepts and methods for studying the environment of human behavior,'' 1969.

\bibitem{duncan1969nonverbal}
S.~Duncan~Jr, ``Nonverbal communication.'' \emph{Psychological bulletin}, vol.~72, no.~2, p. 118, 1969.

\bibitem{wiener1972nonverbal}
M.~Wiener, S.~Devoe, S.~Rubinow, and J.~Geller, ``Nonverbal behavior and nonverbal communication.'' \emph{Psychological review}, vol.~79, no.~3, p. 185, 1972.

\bibitem{lafrance1978cultural}
M.~LaFrance and C.~Mayo, ``Cultural aspects of nonverbal communication,'' \emph{International Journal of Intercultural Relations}, vol.~2, no.~1, pp. 71--89, 1978.

\bibitem{ziemke2007body}
T.~Ziemke, J.~Zlatev, and R.~M. Frank, \emph{Body, language, and mind}.\hskip 1em plus 0.5em minus 0.4em\relax Walter de Gruyter, 2007, vol.~35.

\bibitem{beck2010towards}
A.~Beck, L.~Ca{\~n}amero, and K.~A. Bard, ``Towards an affect space for robots to display emotional body language,'' in \emph{19th International symposium in robot and human interactive communication}.\hskip 1em plus 0.5em minus 0.4em\relax IEEE, 2010, pp. 464--469.

\bibitem{saleem2021real}
S.~M. Saleem, S.~R. Zeebaree, and M.~B. Abdulrazzaq, ``Real-life dynamic facial expression recognition: a review,'' in \emph{Journal of Physics: Conference Series}, vol. 1963, no.~1.\hskip 1em plus 0.5em minus 0.4em\relax IOP Publishing, 2021, p. 012010.

\bibitem{guo2012dynamic}
Y.~Guo, G.~Zhao, and M.~Pietik{\"a}inen, ``Dynamic facial expression recognition using longitudinal facial expression atlases,'' in \emph{Computer Vision--ECCV 2012: 12th European Conference on Computer Vision, Florence, Italy, October 7-13, 2012, Proceedings, Part II 12}.\hskip 1em plus 0.5em minus 0.4em\relax Springer, 2012, pp. 631--644.

\bibitem{fang2014facial}
H.~Fang, N.~Mac~Parthal{\'a}in, A.~J. Aubrey, G.~K. Tam, R.~Borgo, P.~L. Rosin, P.~W. Grant, D.~Marshall, and M.~Chen, ``Facial expression recognition in dynamic sequences: An integrated approach,'' \emph{Pattern Recognition}, vol.~47, no.~3, pp. 1271--1281, 2014.

\bibitem{montirosso2010development}
R.~Montirosso, M.~Peverelli, E.~Frigerio, M.~Crespi, and R.~Borgatti, ``The development of dynamic facial expression recognition at different intensities in 4-to 18-year-olds,'' \emph{Social Development}, vol.~19, no.~1, pp. 71--92, 2010.

\bibitem{liu2014learning}
M.~Liu, S.~Shan, R.~Wang, and X.~Chen, ``Learning expressionlets on spatio-temporal manifold for dynamic facial expression recognition,'' in \emph{Proceedings of the IEEE conference on computer vision and pattern recognition}, 2014, pp. 1749--1756.

\bibitem{tao2023freq}
\BIBentryALTinterwordspacing
Z.~Tao, Y.~Wang, Z.~Chen, B.~Wang, S.~Yan, K.~Jiang, S.~Gao, and W.~Zhang, ``Freq-hd: An interpretable frequency-based high-dynamics affective clip selection method for in-the-wild facial expression recognition in videos,'' in \emph{Proceedings of the 31st ACM International Conference on Multimedia}, ser. MM '23.\hskip 1em plus 0.5em minus 0.4em\relax New York, NY, USA: Association for Computing Machinery, 2023, p. 843–852. [Online]. Available: \url{https://doi.org/10.1145/3581783.3611972}
\BIBentrySTDinterwordspacing

\bibitem{dosovitskiy2020image}
A.~Dosovitskiy, L.~Beyer, A.~Kolesnikov, D.~Weissenborn, X.~Zhai, T.~Unterthiner, M.~Dehghani, M.~Minderer, G.~Heigold, S.~Gelly \emph{et~al.}, ``An image is worth 16x16 words: Transformers for image recognition at scale,'' \emph{arXiv preprint arXiv:2010.11929}, 2020.

\bibitem{vaswani2017attention}
A.~Vaswani, N.~Shazeer, N.~Parmar, J.~Uszkoreit, L.~Jones, A.~N. Gomez, {\L}.~Kaiser, and I.~Polosukhin, ``Attention is all you need,'' \emph{Advances in neural information processing systems}, vol.~30, 2017.

\bibitem{radford2021learning}
A.~Radford, J.~W. Kim, C.~Hallacy, A.~Ramesh, G.~Goh, S.~Agarwal, G.~Sastry, A.~Askell, P.~Mishkin, J.~Clark \emph{et~al.}, ``Learning transferable visual models from natural language supervision,'' in \emph{International conference on machine learning}.\hskip 1em plus 0.5em minus 0.4em\relax PMLR, 2021, pp. 8748--8763.

\bibitem{li2023cliper}
H.~Li, H.~Niu, Z.~Zhu, and F.~Zhao, ``Cliper: A unified vision-language framework for in-the-wild facial expression recognition,'' \emph{arXiv preprint arXiv:2303.00193}, 2023.

\bibitem{tao20243}
Z.~Tao, Y.~Wang, J.~Lin, H.~Wang, X.~Mai, J.~Yu, X.~Tong, Z.~Zhou, S.~Yan, Q.~Zhao \emph{et~al.}, ``A3lign-dfer: Pioneering comprehensive dynamic affective alignment for dynamic facial expression recognition with clip,'' \emph{arXiv preprint arXiv:2403.04294}, 2024.

\bibitem{lucey2010extended}
P.~Lucey, J.~F. Cohn, T.~Kanade, J.~Saragih, Z.~Ambadar, and I.~Matthews, ``The extended cohn-kanade dataset (ck+): A complete dataset for action unit and emotion-specified expression,'' in \emph{2010 ieee computer society conference on computer vision and pattern recognition-workshops}.\hskip 1em plus 0.5em minus 0.4em\relax IEEE, 2010, pp. 94--101.

\bibitem{gera2021landmark}
D.~Gera and S.~Balasubramanian, ``Landmark guidance independent spatio-channel attention and complementary context information based facial expression recognition,'' \emph{Pattern Recognition Letters}, vol. 145, pp. 58--66, 2021.

\bibitem{ekenel2012benchmarking}
H.~K. Ekenel, ``Benchmarking facial image analysis technologies (befit),'' in \emph{2012 3rd International Conference on Image Processing Theory, Tools and Applications (IPTA)}.\hskip 1em plus 0.5em minus 0.4em\relax IEEE, 2012, pp. 15--15.

\bibitem{valstar2010induced}
M.~Valstar, M.~Pantic \emph{et~al.}, ``Induced disgust, happiness and surprise: an addition to the mmi facial expression database,'' in \emph{Proc. 3rd Intern. Workshop on EMOTION (satellite of LREC): Corpora for Research on Emotion and Affect}, vol.~10.\hskip 1em plus 0.5em minus 0.4em\relax Paris, France., 2010, p.~65.

\bibitem{jiang2020dfew}
X.~Jiang, Y.~Zong, W.~Zheng, C.~Tang, W.~Xia, C.~Lu, and J.~Liu, ``Dfew: A large-scale database for recognizing dynamic facial expressions in the wild,'' in \emph{Proceedings of the 28th ACM international conference on multimedia}, 2020, pp. 2881--2889.

\bibitem{wang2022ferv39k}
Y.~Wang, Y.~Sun, Y.~Huang, Z.~Liu, S.~Gao, W.~Zhang, W.~Ge, and W.~Zhang, ``Ferv39k: A large-scale multi-scene dataset for facial expression recognition in videos,'' in \emph{Proceedings of the IEEE/CVF conference on computer vision and pattern recognition}, 2022, pp. 20\,922--20\,931.

\bibitem{guo2016dynamic}
Y.~Guo, G.~Zhao, and M.~Pietik{\"a}inen, ``Dynamic facial expression recognition with atlas construction and sparse representation,'' \emph{IEEE Transactions on Image Processing}, vol.~25, no.~5, pp. 1977--1992, 2016.

\bibitem{zhou2022conditional}
K.~Zhou, J.~Yang, C.~C. Loy, and Z.~Liu, ``Conditional prompt learning for vision-language models,'' in \emph{Proceedings of the IEEE/CVF Conference on Computer Vision and Pattern Recognition}, 2022, pp. 16\,816--16\,825.

\bibitem{tran2015learning}
D.~Tran, L.~Bourdev, R.~Fergus, L.~Torresani, and M.~Paluri, ``Learning spatiotemporal features with 3d convolutional networks,'' in \emph{Proceedings of the IEEE international conference on computer vision}, 2015, pp. 4489--4497.

\bibitem{carreira2017quo}
J.~Carreira and A.~Zisserman, ``Quo vadis, action recognition? a new model and the kinetics dataset,'' in \emph{proceedings of the IEEE Conference on Computer Vision and Pattern Recognition}, 2017, pp. 6299--6308.

\bibitem{qiu2017learning}
Z.~Qiu, T.~Yao, and T.~Mei, ``Learning spatio-temporal representation with pseudo-3d residual networks,'' in \emph{proceedings of the IEEE International Conference on Computer Vision}, 2017, pp. 5533--5541.

\bibitem{hara2018can}
K.~Hara, H.~Kataoka, and Y.~Satoh, ``Can spatiotemporal 3d cnns retrace the history of 2d cnns and imagenet?'' in \emph{Proceedings of the IEEE conference on Computer Vision and Pattern Recognition}, 2018, pp. 6546--6555.

\bibitem{he2016deep}
K.~He, X.~Zhang, S.~Ren, and J.~Sun, ``Deep residual learning for image recognition,'' in \emph{Proceedings of the IEEE conference on computer vision and pattern recognition}, 2016, pp. 770--778.

\bibitem{zhao2023prompting}
Z.~Zhao and I.~Patras, ``Prompting visual-language models for dynamic facial expression recognition,'' \emph{arXiv preprint arXiv:2308.13382}, 2023.

\bibitem{foteinopoulou2023emoclip}
N.~M. Foteinopoulou and I.~Patras, ``Emoclip: A vision-language method for zero-shot video facial expression recognition,'' \emph{arXiv preprint arXiv:2310.16640}, 2023.

\bibitem{ma2023logo}
F.~Ma, B.~Sun, and S.~Li, ``Logo-former: Local-global spatio-temporal transformer for dynamic facial expression recognition,'' in \emph{ICASSP 2023-2023 IEEE International Conference on Acoustics, Speech and Signal Processing (ICASSP)}.\hskip 1em plus 0.5em minus 0.4em\relax IEEE, 2023, pp. 1--5.

\bibitem{li2022nr}
H.~Li, M.~Sui, Z.~Zhu \emph{et~al.}, ``Nr-dfernet: Noise-robust network for dynamic facial expression recognition,'' \emph{arXiv preprint arXiv:2206.04975}, 2022.

\bibitem{li2023scaling}
Y.~Li, H.~Fan, R.~Hu, C.~Feichtenhofer, and K.~He, ``Scaling language-image pre-training via masking,'' in \emph{Proceedings of the IEEE/CVF Conference on Computer Vision and Pattern Recognition}, 2023, pp. 23\,390--23\,400.

\bibitem{wang2022dpcnet}
Y.~Wang, Y.~Sun, W.~Song, S.~Gao, Y.~Huang, Z.~Chen, W.~Ge, and W.~Zhang, ``Dpcnet: Dual path multi-excitation collaborative network for facial expression representation learning in videos,'' in \emph{Proceedings of the 30th ACM International Conference on Multimedia}, 2022, pp. 101--110.

\bibitem{li2023intensity}
H.~Li, H.~Niu, Z.~Zhu, and F.~Zhao, ``Intensity-aware loss for dynamic facial expression recognition in the wild,'' in \emph{Proceedings of the AAAI Conference on Artificial Intelligence}, vol.~37, no.~1, 2023, pp. 67--75.

\bibitem{wang2023rethinking}
H.~Wang, B.~Li, S.~Wu, S.~Shen, F.~Liu, S.~Ding, and A.~Zhou, ``Rethinking the learning paradigm for dynamic facial expression recognition,'' in \emph{Proceedings of the IEEE/CVF Conference on Computer Vision and Pattern Recognition}, 2023, pp. 17\,958--17\,968.

\bibitem{lee2023frame}
B.~Lee, H.~Shin, B.~Ku, and H.~Ko, ``Frame level emotion guided dynamic facial expression recognition with emotion grouping,'' in \emph{Proceedings of the IEEE/CVF Conference on Computer Vision and Pattern Recognition}, 2023, pp. 5680--5690.

\bibitem{yan2023empower}
S.~Yan, Y.~Wang, X.~Mai, Q.~Zhao, W.~Song, J.~Huang, Z.~Tao, H.~Wang, S.~Gao, and W.~Zhang, ``Empower smart cities with sampling-wise dynamic facial expression recognition via frame-sequence contrastive learning,'' \emph{Computer Communications}, 2023.

\end{thebibliography}

\section{Biography Section}

\vspace{-35pt}
\begin{IEEEbiography}[{\includegraphics[width=1in,height=1.25in,clip,keepaspectratio]{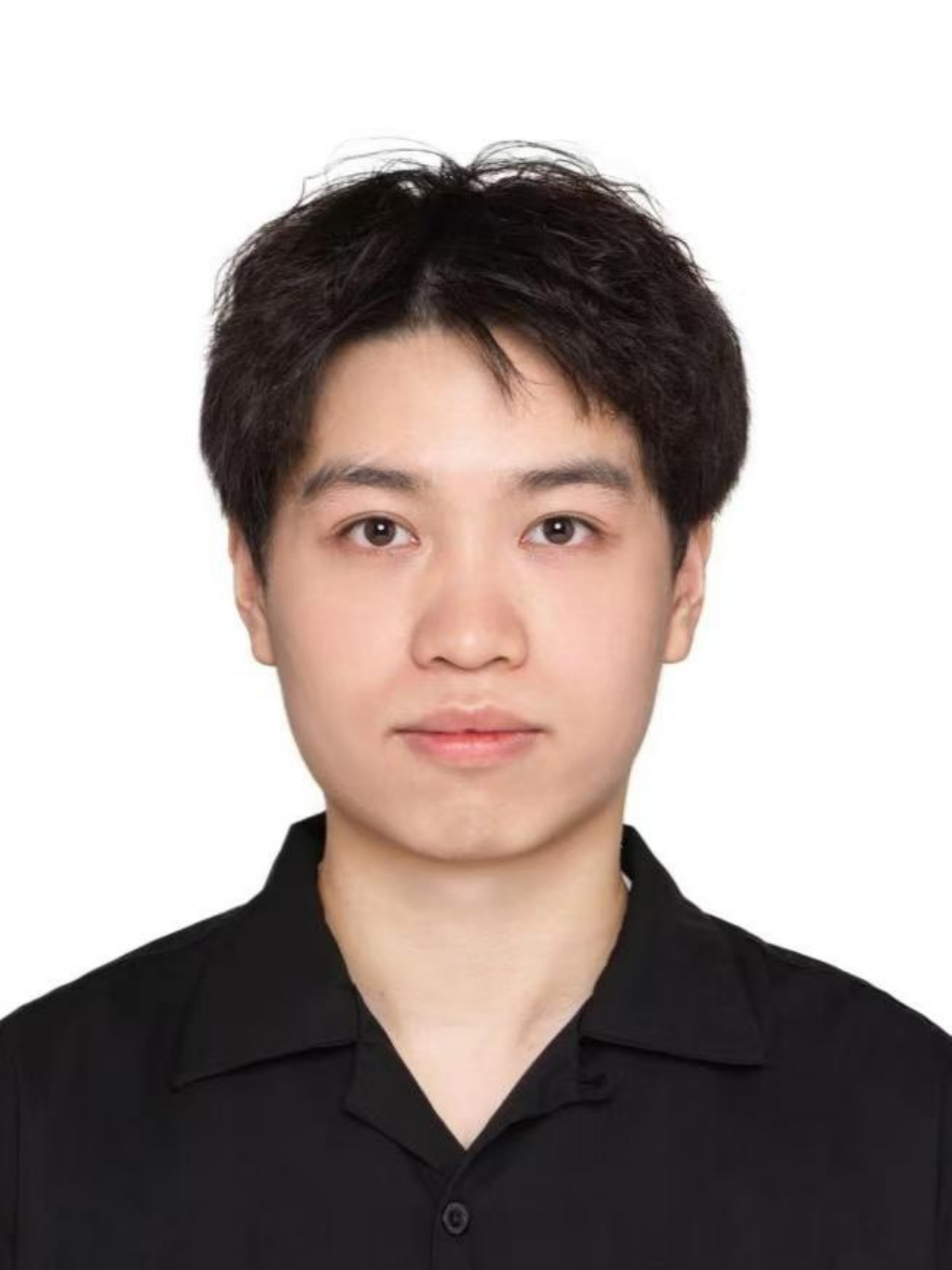}}]{Xinji Mai}
 received the B.E. degree in School of Microelectronics Science and Technology from Sun Yat-sen University, Guangdong, China, in 2023. He is currently a Master’s student in electronics and information in Academy for Engineering and Technology, Fudan University, Shanghai, China.
His current research interests include computer vision and Multimodal large language model.
\end{IEEEbiography}

\vspace{-20pt}
\begin{IEEEbiography}[{\includegraphics[width=1in,height=1.25in,clip,keepaspectratio]{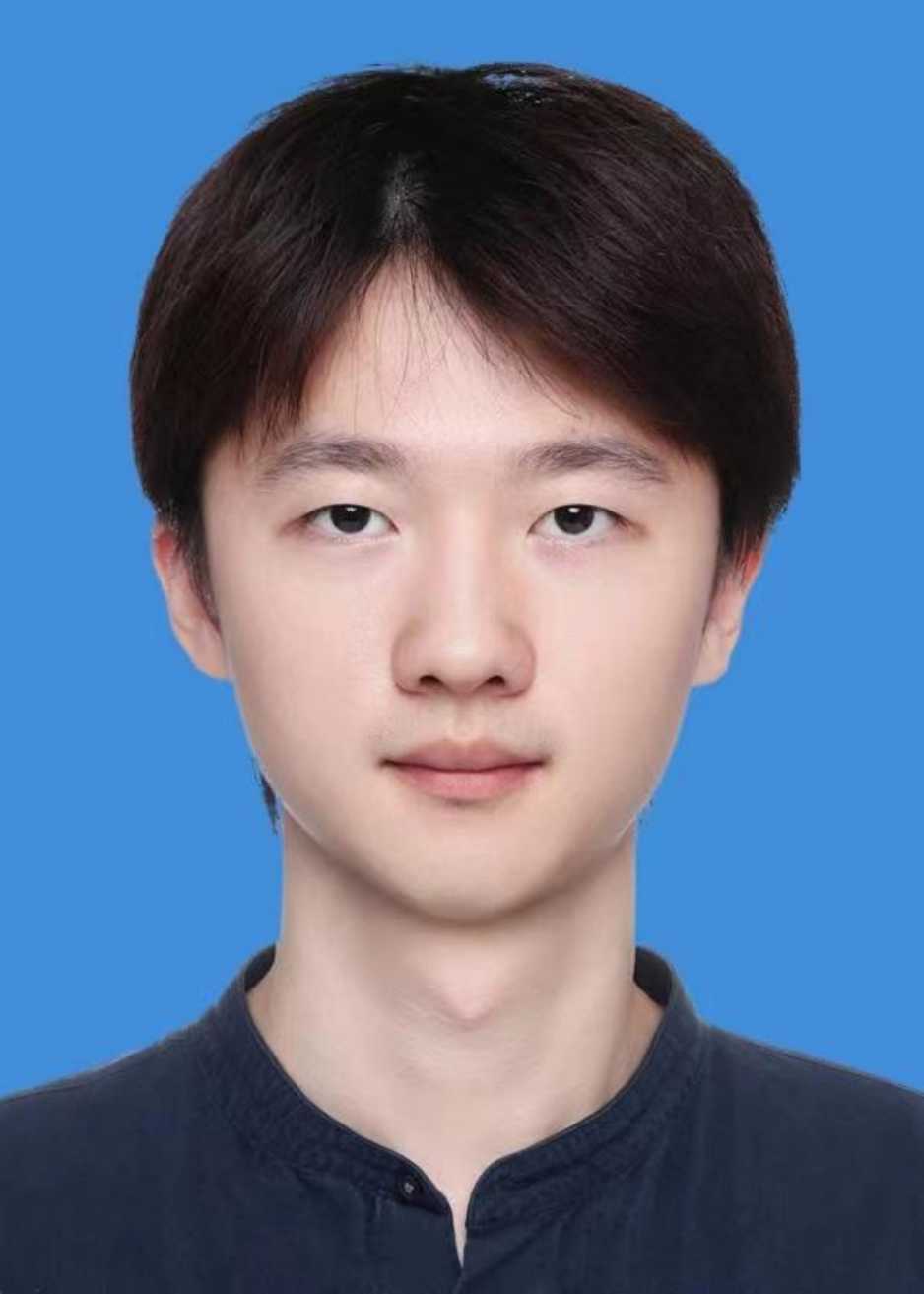}}]{Haoran Wang}
 received the B.E. degree in computer science from Shanghai University, Shanghai, China, in 2023. He is currently working toward the master’s degree
in electronics and information with Academy for
Engineering and Technology, Fudan University,
Shanghai, China.
His research interests include computer vision and affective computing.
\end{IEEEbiography}

\vspace{-20pt}
\begin{IEEEbiography}[{\includegraphics[width=1in,height=1.25in,clip,keepaspectratio]{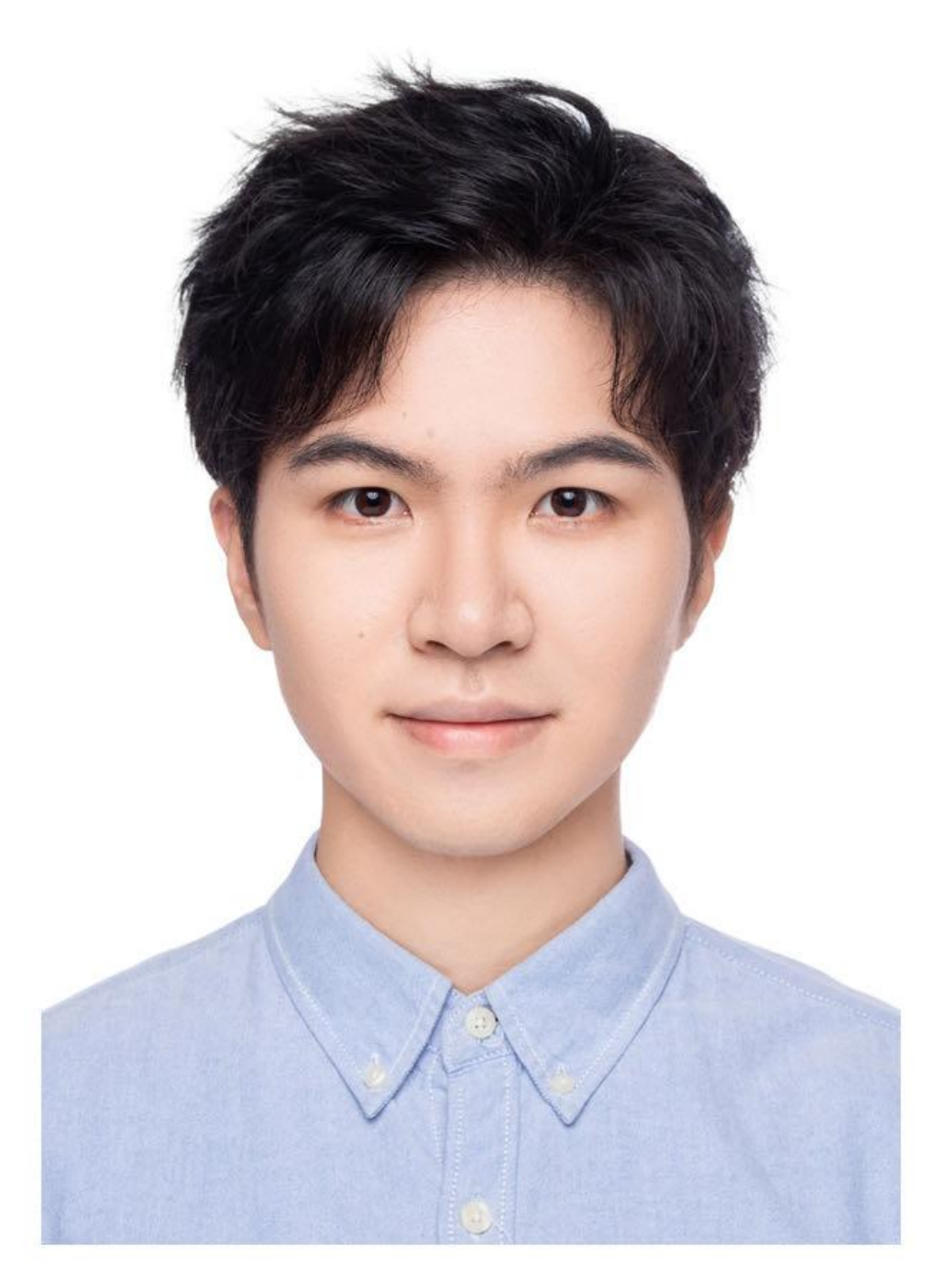}}]{Zeng Tao}
  received the B.E. degree in intelligence science and technology from Fudan University, Shanghai, China, in 2023. He is currently a Master’s student in computer science and technology in Academy for Engineering and Technology, Fudan University, Shanghai, China.
His current research interests include computer vision and generation.
\end{IEEEbiography}

\vspace{-20pt}
\begin{IEEEbiography}[{\includegraphics[width=1in,height=1.25in,clip,keepaspectratio]{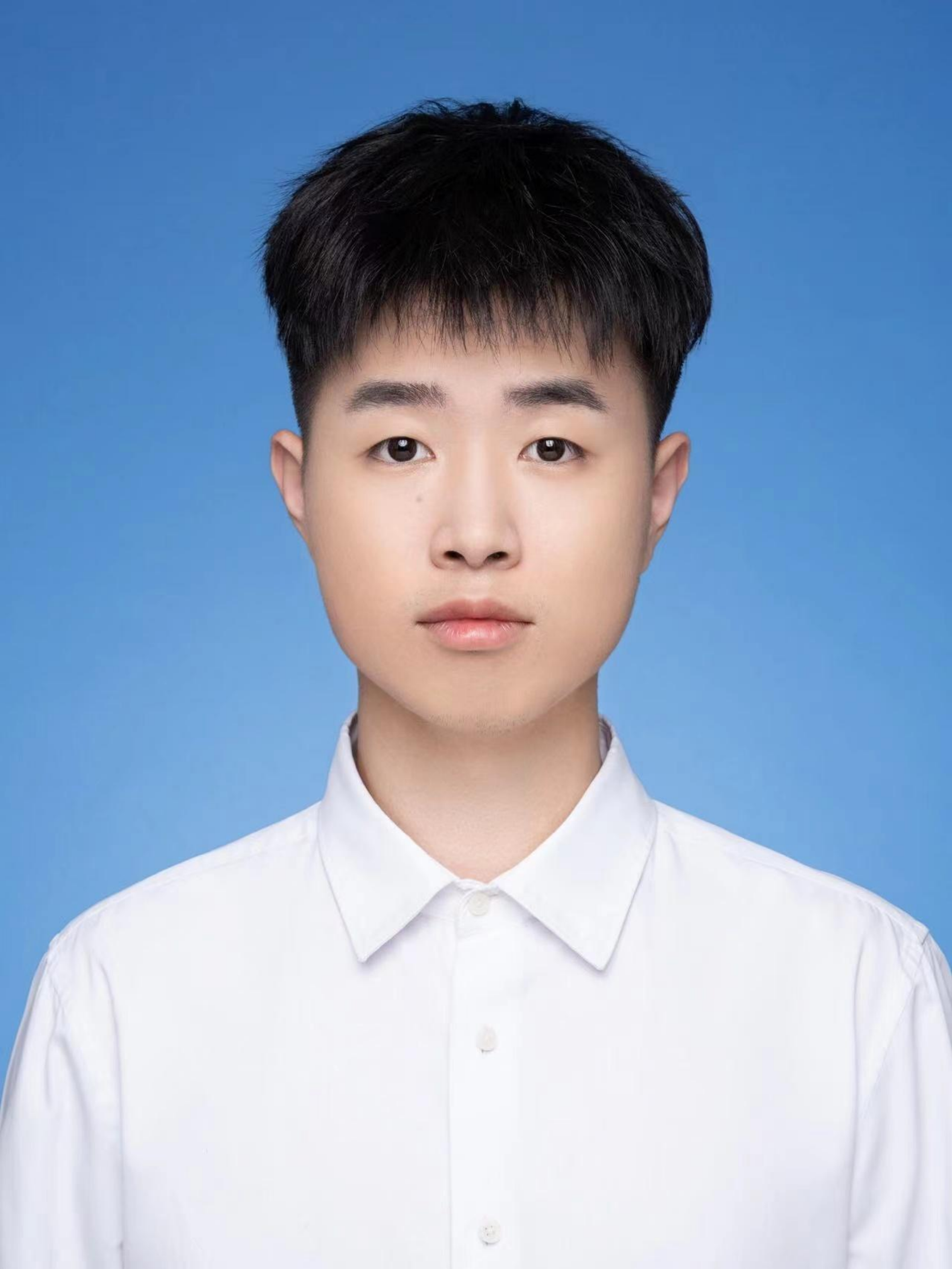}}]{Junxiong Lin}
  received the B.S. degree in information and computational science from Shenzhen University, Shenzhen, China, in 2022. He is currently working toward the master’s degree in electronics and information with Academy for Engineering and Technology, Fudan University, Shanghai, China.
His research interests include deep learning and computer vision.
\end{IEEEbiography}

\vspace{-20pt}
\begin{IEEEbiography}[{\includegraphics[width=1in,height=1.25in,clip,keepaspectratio]{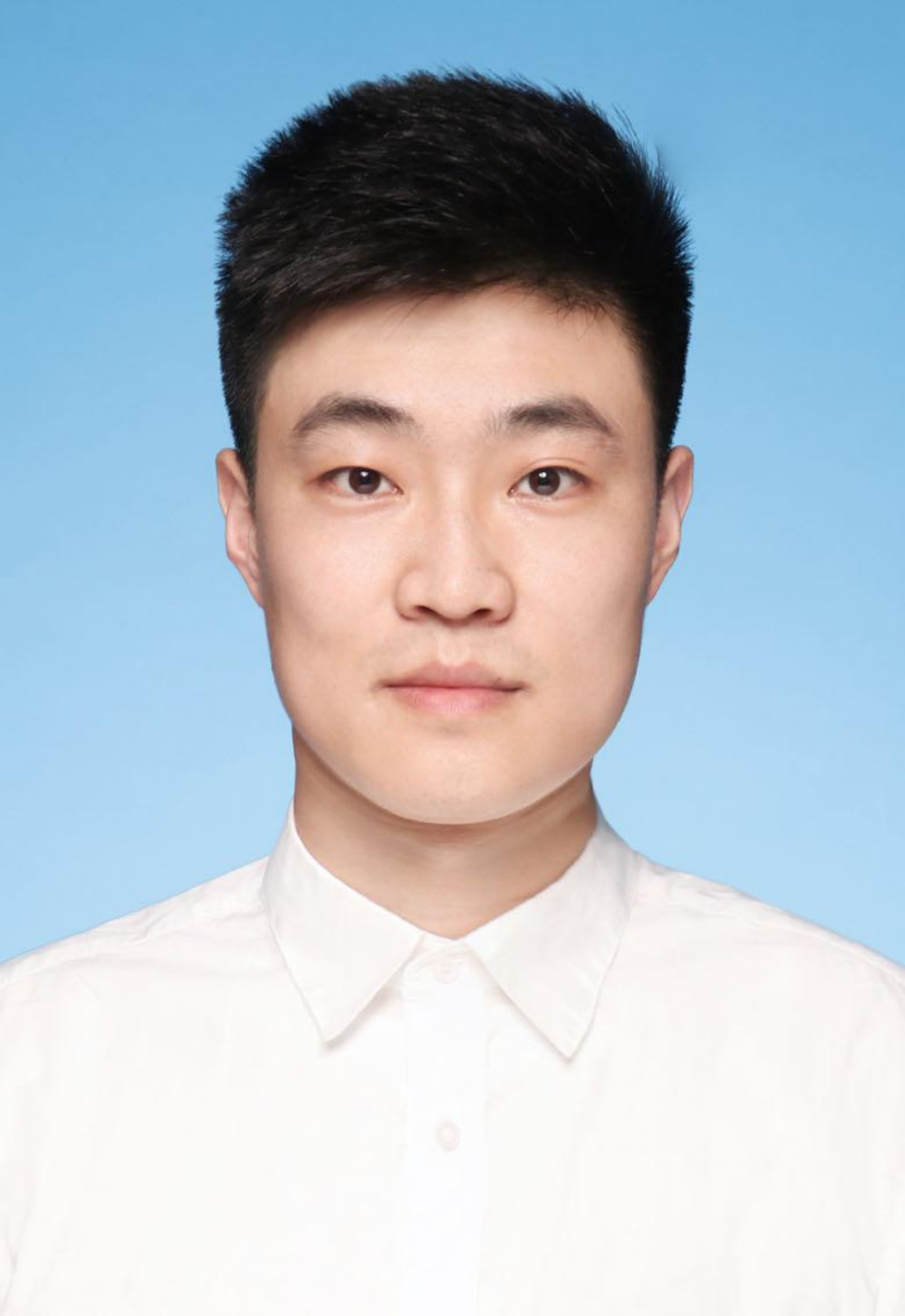}}]{Shaoqi Yan}
  received the M.S. degree in software engineering from Fudan University, Shanghai, China, in 2019. He is currently working toward the Ph.D. degree in electronics and information with the Academy for Engineering and Technology, Fudan University. His research interests include computer vision, affective computing, and robot intelligence.
\end{IEEEbiography}

\vspace{-20pt}
\begin{IEEEbiography}[{\includegraphics[width=1in,height=1.25in,clip,keepaspectratio]{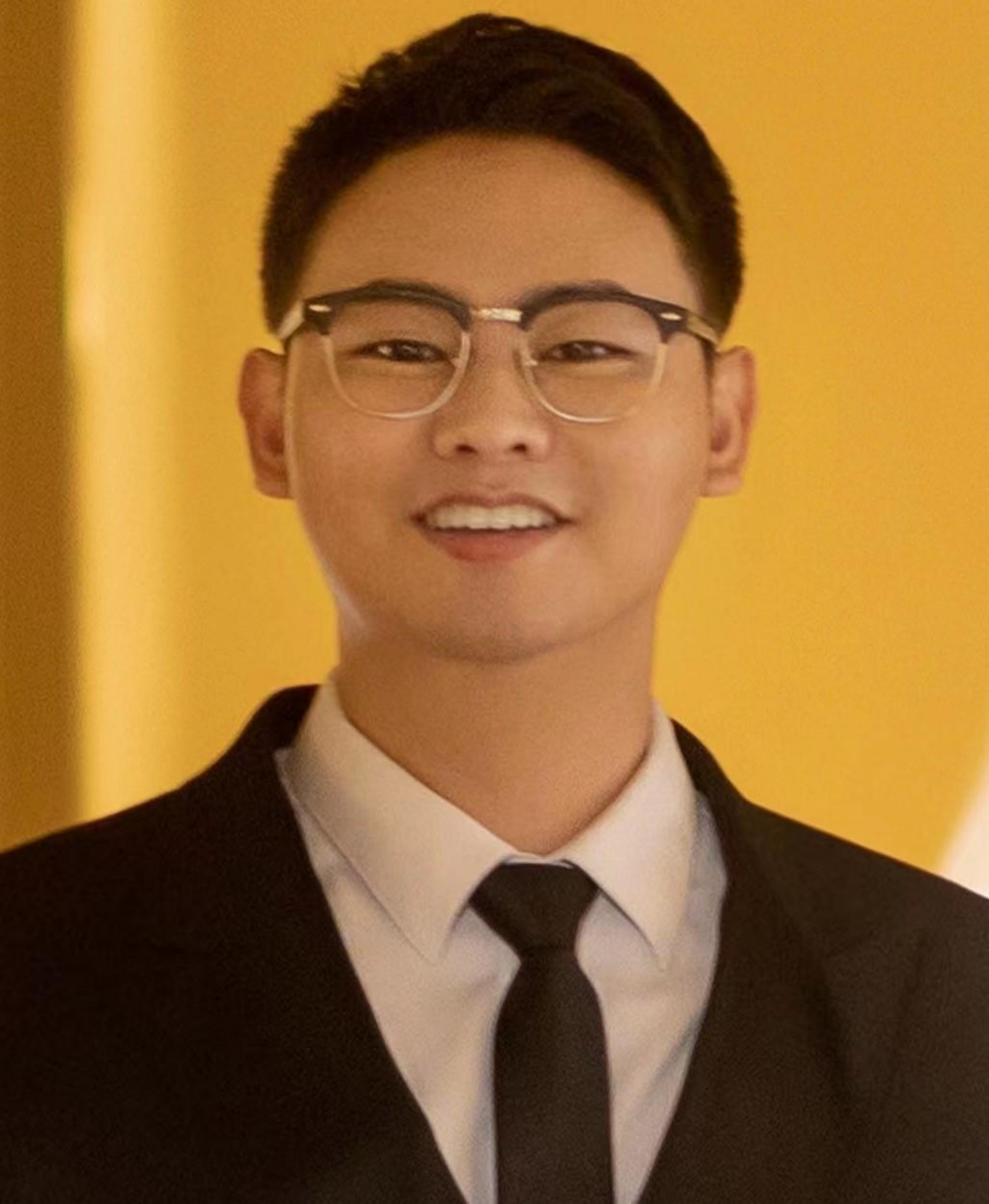}}]{Yan Wang}
  received the Ph.D. degree in Academy for Engineering and Technology from Fudan University, Shanghai, China, in 2023. He is currently a Postdoctoral Fellow at the Institute of Intelligent Robotics, Fudan University. He has authored or coauthored more than 20 papers in prestigious journals and conferences. His research interests include intelligent affective Robot, product appearance detecion under complex working conditions, and AI for Science.  He is a PC member for top conferences, such as CVPR, ICCV, ECCV, NeurIPS, ACM MM and AAAI.
\end{IEEEbiography}

\vspace{-20pt}
\begin{IEEEbiography}[{\includegraphics[width=1in,height=1.25in,clip,keepaspectratio]{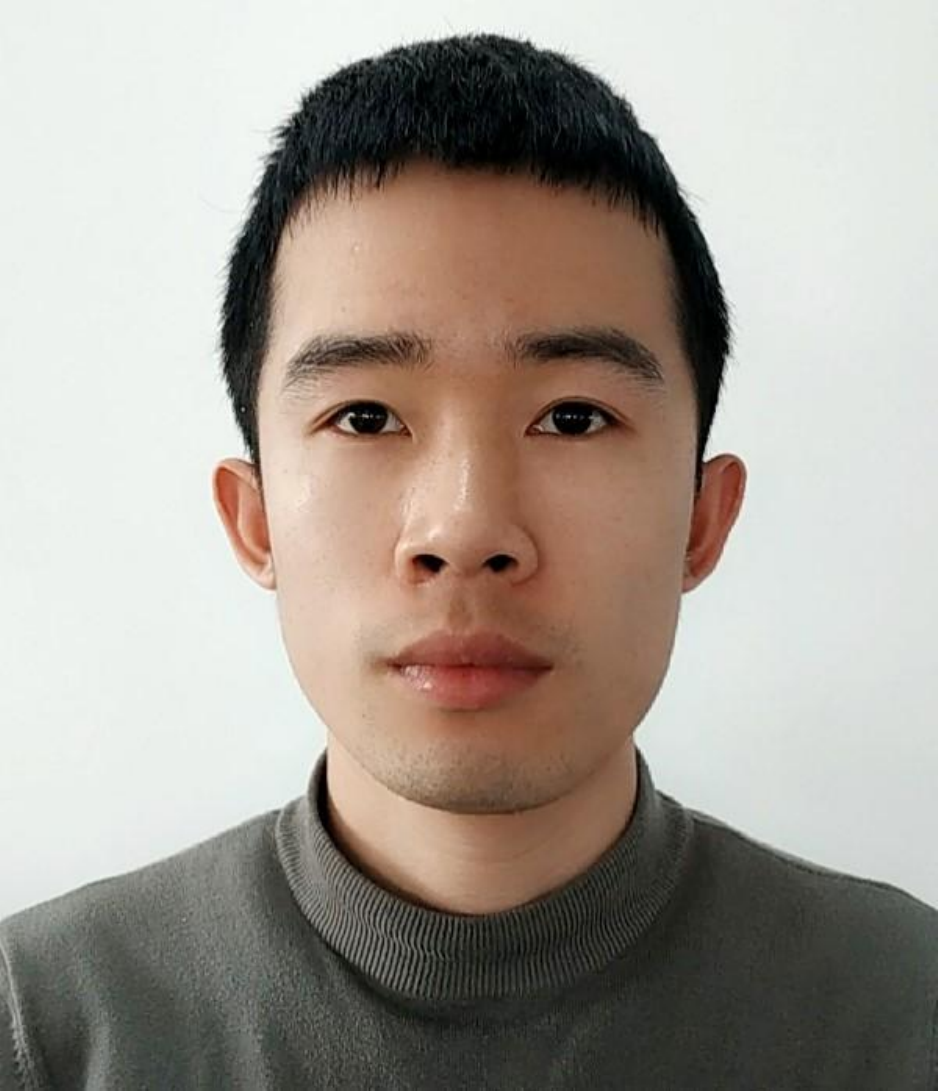}}]{Jing Liu}
  (Member, IEEE) received the B.Eng. and M.S. degrees in computer science and technology, and software engineering from the University of Shanghai for Science and Technology, Shanghai, China, in 2014 and 2017, respectively, and the Ph.D. degree in electronics and information from Fudan University, Shanghai, China, in 2023.
His research interests include object detection, end-cloud collaboration, domain generalization, and video anomaly detection.
\end{IEEEbiography}

\vspace{-20pt}
\begin{IEEEbiography}[{\includegraphics[width=1in,height=1.25in,clip,keepaspectratio]{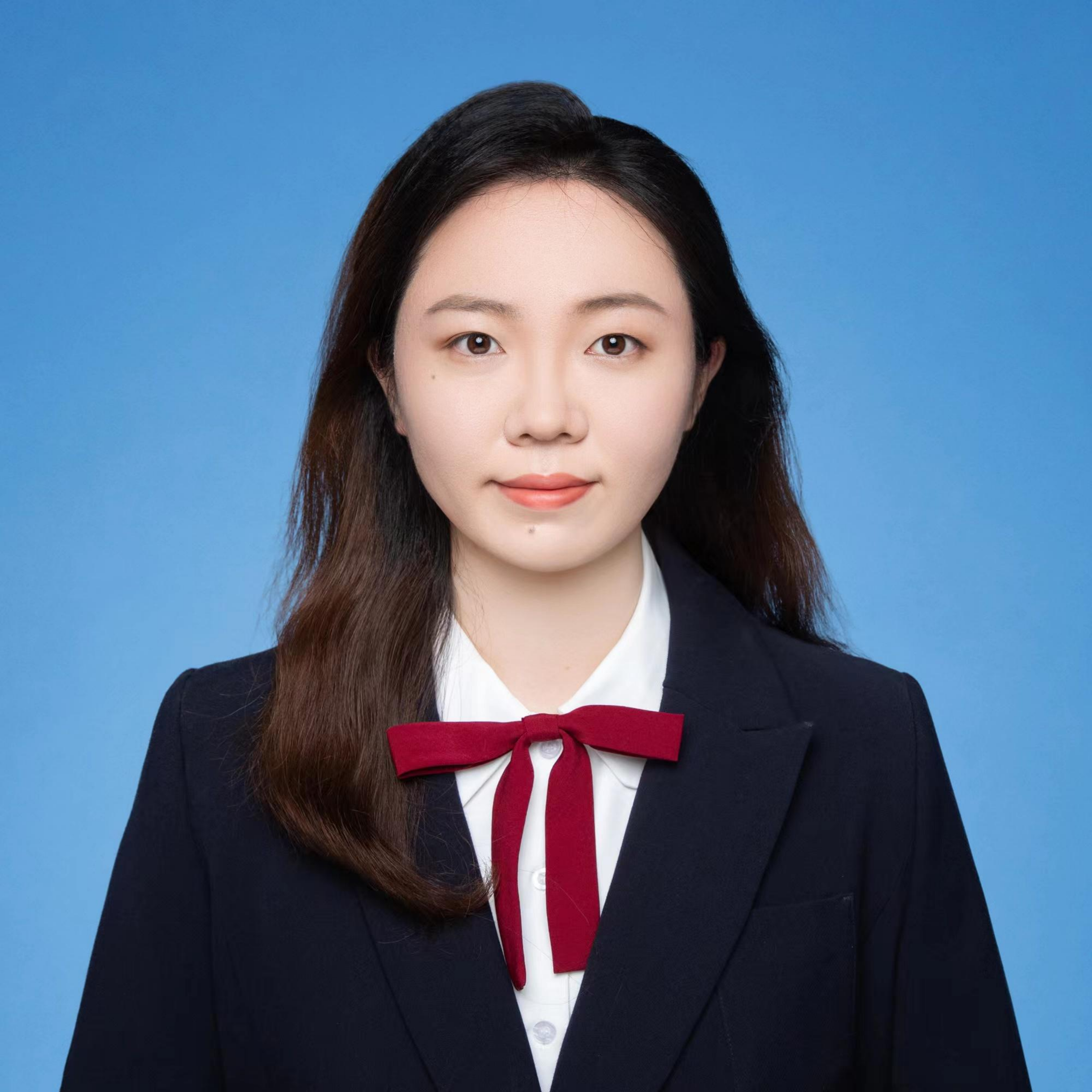}}]{Jiawen Yu}
is currently a Master student at Academy for Engineering \& Technology, Fudan University. She is supervised by Prof. Wenqiang Zhang. She obtained her B.Eng. degree from University of Shanghai for Science and Technology by 2023. Her research interests are in computer vision, and their applications, such as anomaly detection.
\end{IEEEbiography}

\vspace{-20pt}
\begin{IEEEbiography}[{\includegraphics[width=1in,height=1.25in,clip,keepaspectratio]{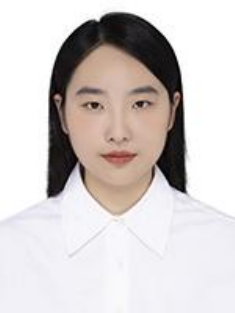}}]{Xuan Tong}
  is currently a Master student at Academy for Engineering \& Technology, Fudan University. She is supervised by Prof. Wenqiang Zhang. She obtained her B.Eng. degree from Chongqing University by 2023. Her research interests are in computer vision, and their applications, such as anomaly detection.
\end{IEEEbiography}

\vspace{-20pt}
\begin{IEEEbiography}[{\includegraphics[width=1in,height=1.25in,clip,keepaspectratio]{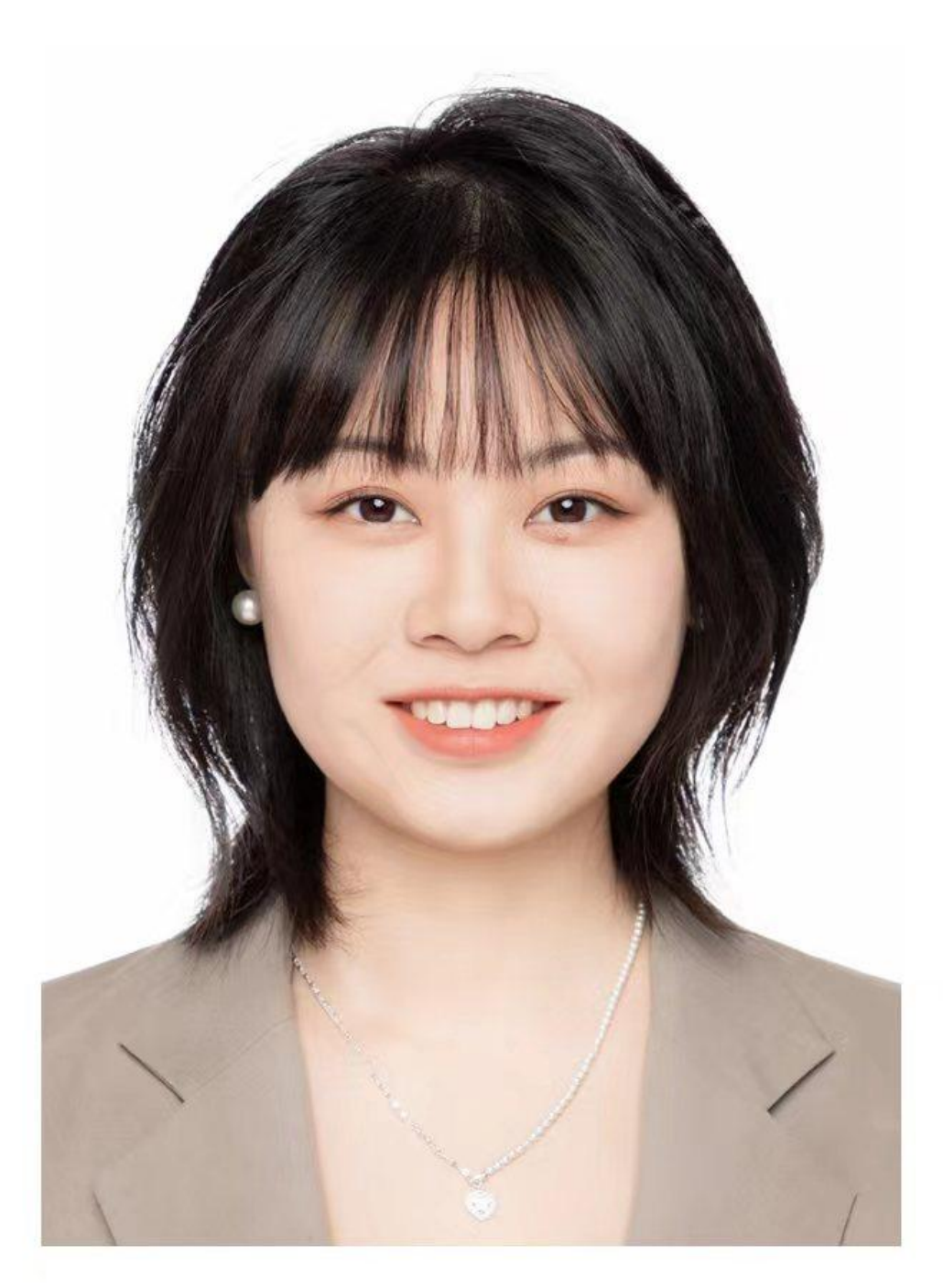}}]{Yating Li}
 received the Bachelor of Management degree in Tourism Academy from Sun Yat-sen University, Guangdong, China, in 2023. She is currently working in BYD Co., Ltd.
Her current research interests include Artificial Intelligence and Art.

\end{IEEEbiography}

\vspace{-20pt}
\begin{IEEEbiography}[{\includegraphics[width=1in,height=1.25in,clip,keepaspectratio]{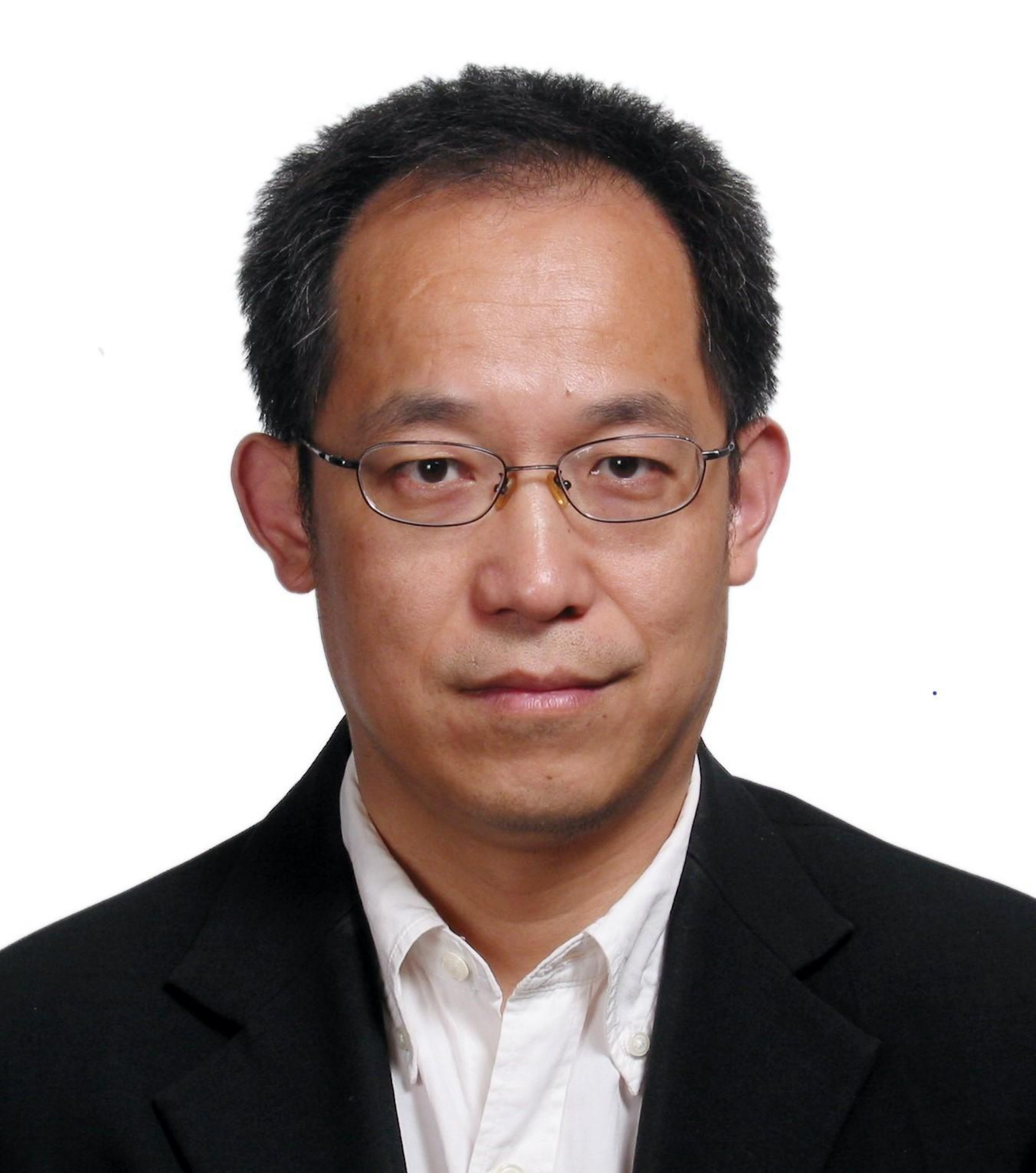}}]{Wenqiang Zhang}
  received the Ph.D. degree from Shanghai Jiao Tong University, China. He is currently a Professor with the School of Computer Science, Fudan University. His current research interests include computer vision, and robot intelligence. He has undertaken several major program projects from National Key R\&D Program of China , and the National Natural Science Foundation of China (NSFC).

\end{IEEEbiography}

\vfill

\end{document}